\documentclass{article}

\PassOptionsToPackage{numbers}{natbib}
\usepackage[final]{neurips_2020}

\usepackage[utf8]{inputenc} 
\usepackage[T1]{fontenc}    
\usepackage{hyperref}       
\usepackage{url}            
\usepackage{booktabs}       
\usepackage{amsfonts}       
\usepackage{nicefrac}       
\usepackage{microtype}      

\usepackage{enumitem}
\usepackage{lipsum}
\usepackage{amsmath}
\usepackage{amssymb}
\usepackage{amsthm}
\usepackage{bm}
\usepackage{dsfont}
\usepackage{graphicx}
\usepackage{wrapfig}
\usepackage{subcaption}
\usepackage{adjustbox}

\usepackage{algorithm, algcompatible}
\usepackage{multicol,changepage}
\usepackage[noend]{algpseudocode}

\usepackage[dvipsnames]{xcolor}
\newif\ifcomments
\commentstrue
\ifcomments
  \newcommand{\colornote}[3]{{\color{#1}\bf{#2: #3}\normalfont}}
\else
  \newcommand{\colornote}[3]{}
\fi

\def\independenT#1#2{\mathrel{\rlap{$#1#2$}\mkern2mu{#1#2}}}
\newcommand{\mybm}[1]{\scalebox{0.9}[1]{$\bm{#1}$}}
\newcommand\independent{\protect\mathpalette{\protect\independenT}{\perp}}
\newcommand{\indep}{\independent}
\newcommand{\given}{\,|\,}
\newcommand{\DO}{\textrm{do}}
\newcommand{\s}{\mathbf{s}}
\newcommand{\R}{\mathbb{R}}

\renewcommand{\S}{\mathcal{S}}
\newcommand{\M}{\mathcal{M}}
\renewcommand{\L}{\mathcal{L}}
\newcommand{\A}{\mathcal{A}}

\newcommand{\F}{\mathcal{F}}
\newcommand{\G}{\mathcal{G}}

\newcommand{\W}{\mathbf{W}}

\newcommand{\B}{\mathbf{b}}
\newcommand{\parents}{\textrm{Pa}}

\newcommand{\minimize}{\operatorname{minimize}}

\newcommand{\indicator}[1]{\mathds{1} (#1)}  
\newcommand\restrict[1]{\raisebox{-.2ex}{$|$}_{#1}}
\newcommand\mydots{\ifmmode\ldots\else\makebox[1em][c]{.\hfil.\hfil.}\thinspace\fi}
\newcommand{\localSA}{\mathcal{L}^{(j\indep i)}}

\newtheorem*{assumption}{Assumption}
\newtheorem*{definition}{Definition}
\newtheorem{proposition}{Proposition}
\newtheorem{lemma}{Lemma}
\newtheorem{corollary}{Corollary}
\newtheorem{remark}{Remark}[section]

\makeatletter
\newenvironment{appdxProp}[1]
  {\count@\c@proposition
   \global\c@proposition#1 %
    \global\advance\c@proposition\m@ne
   \proposition}
  {\endproposition
   \global\c@proposition\count@}
\makeatother

\title{Counterfactual Data Augmentation\\ using Locally Factored Dynamics}

\author{%
Silviu Pitis, Elliot Creager, Animesh Garg\\
Department of Computer Science, University of Toronto, Vector Institute\\
\texttt{\{spitis, creager, garg\}@cs.toronto.edu} 
}

\usepackage{subfiles} 

\begin{document}

\maketitle

\begin{abstract}
Many dynamic processes, including common scenarios in robotic control and reinforcement learning (RL), involve a set of interacting subprocesses. Though the subprocesses are not independent, their interactions are often sparse, and the dynamics at any given time step can often be decomposed into \textit{locally independent} causal mechanisms. Such local causal structures can be leveraged to improve the sample efficiency of sequence prediction and off-policy reinforcement learning. We formalize this by introducing local causal models (LCMs), which are induced from a global causal model by conditioning on a subset of the state space. We propose an approach to inferring these structures given an object-oriented state representation, as well as a novel algorithm for Counterfactual Data Augmentation (CoDA). CoDA uses local structures and an experience replay to generate counterfactual experiences that are causally valid in the global model. We find that CoDA significantly improves the performance of RL agents in locally factored tasks, including the batch-constrained and goal-conditioned settings.\footnote{Code available at \url{https://github.com/spitis/mrl}}
\end{abstract}

\section{Introduction}

High-dimensional dynamical systems are often composed of simple subprocesses that affect one another through sparse interaction. 
If the subprocesses \emph{never} interacted, an agent could realize significant gains in sample efficiency by globally factoring the dynamics and modeling each subprocess independently \cite{guestrin2003efficient,hallak2015off}.
In most cases, however, the subprocesses do \emph{eventually} interact and so the prevailing approach is to model the entire process using a monolithic, unfactored model. 
In this paper, we take advantage of the observation that \textit{locally---during the time between their interactions---the subprocesses are causally independent}. By locally factoring dynamic processes in this way, we are able to capture the benefits of factorization even when their subprocesses interact on the global scale. 

\begin{figure}[!t]
	\centering
	\hspace{-0.02\textwidth}
	\includegraphics[width=0.98\textwidth]{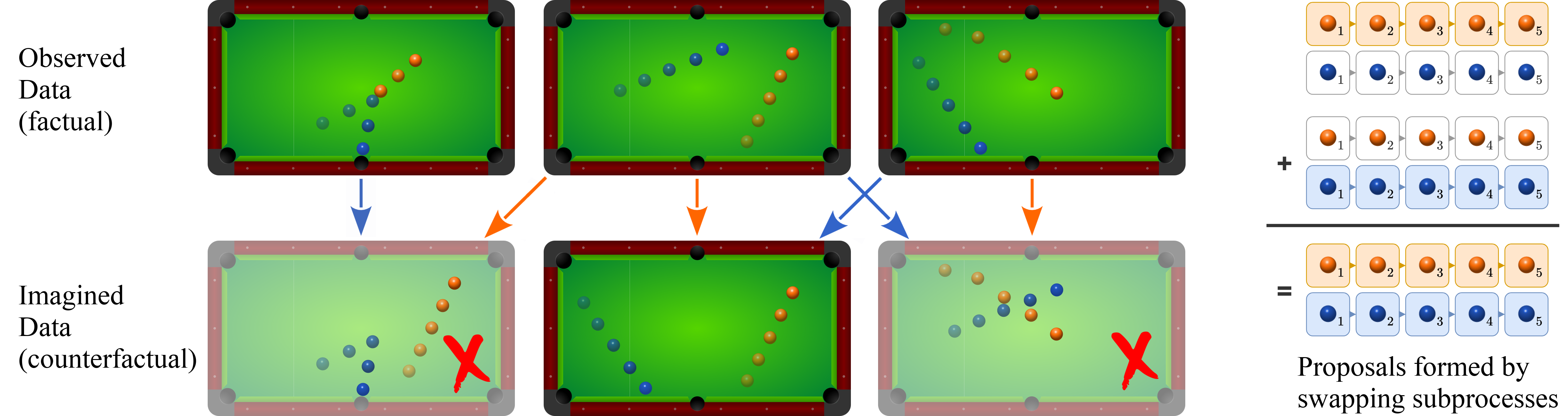}
	\hspace{0.02\textwidth}
	\caption{\textbf{Counterfactual Data Augmentation (CoDA)}. Given 3 factual samples, knowledge of the local causal structure lets us mix and match factored subprocesses to form counterfactual samples. The first proposal is rejected because one of its factual sources (the blue ball) is not locally factored.	The third proposal is rejected because it is not itself factored. The second proposal is accepted, and can be used as additional training data for a reinforcement learning agent.
	}
	\label{fig_coda_intro_diagram}
\end{figure}

Consider a game of billiards, where each ball can be viewed as a separate physical subprocess.
Predicting the opening break is difficult because all balls are mechanically coupled by their initial placement.
Indeed, a dynamics model with dense coupling amongst balls may seem sensible when considering the expected outcomes over the course of the game, as each ball has a non-zero chance of colliding with the others. But at any given timestep, interactions between balls are usually sparse.

One way to take advantage of sparse interactions between otherwise disentangled entities is to use a structured state representation together with a graph neural network or other message passing transition model that captures the local interactions \cite{goyal2019recurrent, kipf2019contrastive}. When it is tractable to do so, such architectures can be used to model the world dynamics directly, producing transferable, task-agnostic models. In many cases, however, the underlying processes are difficult to model precisely, and model-free \cite{laskin2020reinforcement,wang2019benchmarking} or task-oriented model-based  \cite{farahmand2017value,oh2017value} approaches are less biased and exhibit superior performance. In this paper we argue that \emph{knowledge of whether or not local interactions occur is useful in and of itself}, and can be used to generate causally-valid counterfactual data even in absence of a forward dynamics model. 
In fact, if two trajectories have the same local factorization in their transition dynamics, then under mild conditions we can produce new counterfactually plausible data using our proposed \textbf{Counterfactual Data Augmentation (CoDA)} technique, wherein factorized subspaces of observed trajectory pairs are swapped (Figure \ref{fig_coda_intro_diagram}). This lets us sample from a
counterfactual data distribution by stitching together subsamples from observed transitions. Since CoDA acts only on the agent's training data, it is compatible with any agent architecture (including unfactored ones). 

In the remainder of this paper, we formalize this data augmentation strategy and discuss how it can improve performance of model-free RL agents in locally factored tasks. 

Our main contributions are:
\begin{enumerate}[leftmargin=*]
\item We define local causal models (LCMs), which are induced from a global model by conditioning on a subset of the state space, and show how local structure can simplify counterfactual reasoning.
\item We introduce CoDA as a generalized data augmentation strategy that is able to leverage local factorizations to manufacture unseen, yet causally valid, samples of the environment dynamics.
We show that goal relabeling \cite{kaelbling1993learning,andrychowicz2017hindsight} and visual augmentation \cite{andrychowicz2020learning,laskin2020reinforcement} are instances of CoDA that use global independence relations
and we propose a locally conditioned variant of CoDA that swaps independent subprocesses to form counterfactual experiences (Figure \ref{fig_coda_intro_diagram}).
\item Using an attention-based method for discovering local causal structure in a disentangled state space, we show that our CoDA algorithm significantly improves the sample efficiency in standard, batch-constrained, and goal-conditioned reinforcement learning settings. 
\end{enumerate}

\section{
Local Causality in MDPs
}\label{section_locally_factored}

\subsection{Preliminaries and Problem Setup} \label{subsection_prelim}
The basic model for decision making in a controlled dynamic process is a Markov Decision Process (MDP), described by tuple $\langle \S, \A, P, R, \gamma \rangle$ consisting of the state space, action space, transition function, reward function, and discount factor, respectively \cite{puterman2014markov,sutton2018reinforcement}. Note that MDPs generalize uncontrolled Markov processes (set $A = \emptyset$), so that our work applies also to sequential prediction. 
We denote individual states and actions using lowercase $s \in \S$ and $a \in \A$, and variables using the uppercase $S$ and $A$ (e.g., $s \in \textrm{range}(S) \subseteq \S$). 
A policy $\pi : \S \times \A \to [0, 1]$ defines a probability distribution over the agent's actions at each state, and an agent is typically tasked with learning a parameterized policy $\pi_\theta$ that maximizes value $\mathbb{E}_{P,\pi}\sum_t\gamma^tR(s_t, a_t)$. 

In most non-trivial cases, the state $s \in \S$ can be described as an object hierarchy together with global context. 
For instance, this decomposition will emerge naturally in any simulated process or game that is defined using a high-level programming language (e.g., the commonly used Atari \cite{bellemare2013arcade} or Minecraft \cite{johnson2016malmo} simulators). 
In this paper we consider MDPs with a single, known top-level decomposition of the state space $\S = \S^1 \oplus \S^2 \oplus \dots \oplus \S^n$ for fixed $n$, leaving extensions to hierarchical decomposition and multiple representations \cite{higgins2018scan,esmaeili2019structured}, dynamic factor count $n$ \cite{zaheer2017deep}, and (learned) latent representations \cite{burgess2019monet} to future work. The action space might be similarly decomposed: $\A = \A^1 \oplus \A^2 \oplus \dots \oplus \A^m$. 
Given such state and action decompositions, we may model time slice $(t, t\!+\!1)$ using a structural causal model (SCM) $\M_t = \langle V_t, U_t, \F \rangle$ (\cite{pearl2009causal}, Ch. 7) with directed acyclic graph (DAG) $\G$, where:
\begin{itemize}[leftmargin=0.3in,labelsep=0.1in]
\item $V_t = \{V^i_{t[+1]}\}_{i=0}^{2n+m}\!=\!\{S^1_t \mydots S^n_t\!, A^1_t \mydots A^m_t\!, S^1_{t+1} \mydots S^n_{t+1}\}$ are the nodes (variables) of $\G$.
\item $U_t = \{U^i_{t[+1]}\}_{i=0}^{2n+m}$ is a set of noise variables, one for each $V^i$, determined by the initial state, past actions, and environment stochasticity. We assume that noise variables at time $t+1$ are independent from other noise variables: $U^i_{t+1} \indep U^j_{t[+1]} \forall i,j$. The instance $u = (u^1, u^2, \dots, u^{2n + m})$ of $U_t$ denotes an individual realization of the noise variables. 
\item $\F = \{f^i\}_{i=0}^{2n+m}$ is a set of functions (``structural equations'') that map from $U^i_{t[+1]} \times \parents(V^i_{t[+1]})$ to $V^i_{t[+1]}$, where $\parents(V^i_{t[+1]}) \subset V_t\setminus V^i_{t[+1]}$ are the parents of $V^i_{t[+1]}$ in $\G$; hence each $f^i$ is associated with the set of incoming edges to node $V^i_{t[+1]}$ in $\G$; see, e.g., Figure \ref{fig_local_factorization} (center). 
\end{itemize} 

Note that while $V_t$, $U_t$, and $\M_t$ are indexed by $t$ (their distributions change over time), the structural equations $f^i \in \F$ and causal graph $\G$ represent the global transition function $P$ and apply at all times $t$. To reduce clutter, we drop the subscript $t$ on $V$, $U$, and $\M$ when no confusion can arise. 

Critically, we require the set of edges in $\G$ (and thus the number of inputs to each $f^i$) to be \emph{structurally minimal} (\cite{peters2017elements}, Remark 6.6).

\begin{assumption}[Structural Minimality]
$V^j \in \parents(V^i)$ if and only if there exists some $\{u^i, v^{-ij}\}$ with ${u^i \in \textrm{range}(U^i), v^{-ij} \in \textrm{range}(V\setminus\{V^i, V^j\})}$ and pair ${(v^j_1, v^j_2)}$ with ${v^j_1, v^j_2 \in \textrm{range}(V^j)}$ such that $v^i_1 = f^i(\{u^i, v^{-ij}, v^j_1\}) \not= f^i(\{u^i, v^{-ij}, v^j_2 \}) = v^i_2$.
\end{assumption}

Intuitively, structural minimality says that $V^j$ is a parent of $V^i$ if and only if setting the value of $V^j$ can have a nonzero \emph{direct} effect\footnote{Thus parentage does describe knock-on effects, e.g. $V_1$ on $V_3$ in the Markov chain ${V_1 \rightarrow V_2 \rightarrow V_3}$.} on the child $V^i$ through the structural equation $f^i$. The structurally minimal representation is unique \cite{peters2017elements}.

Given structural minimality, we can think of edges in $\G$ as representing global causal dependence.
The probability distribution of $S^i_{t+1}$ is fully specified by its parents $\parents(S^i_{t+1})$ together with its noise variable $U_i$; that is, we have $P(S^i_{t+1} \given S_t, A_t) = P(S^i_{t+1} \given \parents(S^i_{t+1}))$ so that $S^i_{t+1} \indep V^j\given \parents(S^i_{t+1})$ for all nodes $V^j \not\in  \parents(S^i_{t+1})$. 
We call an MDP with this structure a \textit{factored MDP} \cite{kearns1999efficient}. %
When edges in $\G$ are sparse, factored MDPs admit more efficient solutions than unfactored MDPs \cite{guestrin2003efficient}.

\subsection{Local Causal Models (LCMs)}\label{subsection_locally_factored}
\paragraph{Limitations of Global Models}
Unfortunately, even if states and actions can be cleanly decomposed into several nodes, in most practical scenarios the DAG $\G$ is fully connected (or nearly so): since the $f^i$ apply globally, so too does structural minimality, and edge $(S^i_k, S^j_{k+1})$ at time $k$ is present so long as there is a single instance---at any time $t$, no matter how unlikely---in which $S^i_t$ influences $S^j_{t+1}$. In the words of Andrew Gelman, ``\textit{there are (almost) no true zeros}'' \cite{gelman2011causality}. As a result, the factorized causal model $\M_t$, based on globally factorized dynamics, rarely offers an advantage over a simpler causal model that treats states and actions as monolithic entities (e.g., \cite{buesing2018woulda}).

\begin{figure}[!t]
	\centering
	 \begin{minipage}{0.02\textwidth}
			\hfill
	 \end{minipage}%
    \begin{minipage}{0.16\textwidth}
		\includegraphics[width=\textwidth]{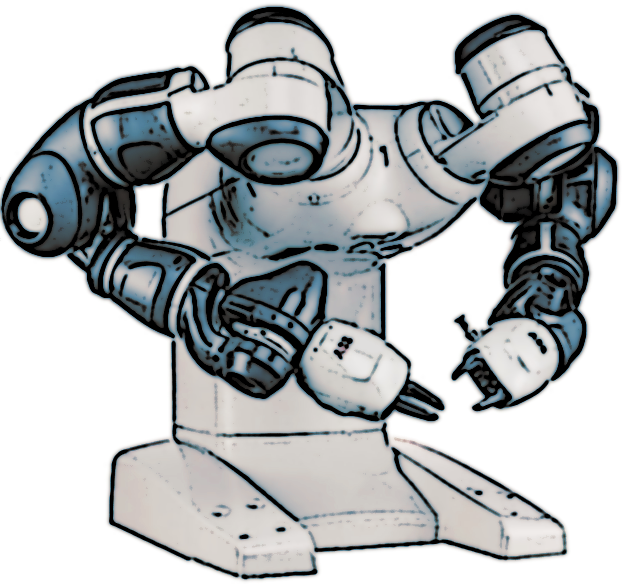}
    \end{minipage}%
    \begin{minipage}{0.05\textwidth}
		\hfill
    \end{minipage}%
    \begin{minipage}{0.77\textwidth}
		\includegraphics[width=\textwidth]{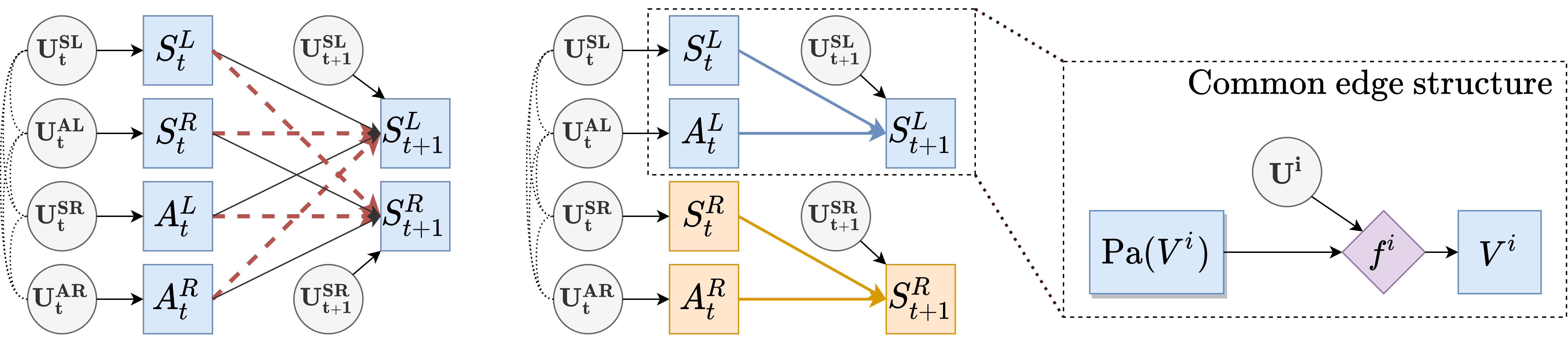}
	\end{minipage}
	\caption{A two-armed robot (\textbf{left}) might be modeled as an MDP whose state and action spaces decompose into left and right subspaces: $\mathcal{S} = \mathcal{S}^L \oplus \mathcal{S}^R, \mathcal{A} = \mathcal{A}^L \oplus \mathcal{A}^R$. Because the arms can touch, the global causal model (\textbf{center left}) between time steps is fully connected, even though left-to-right and right-to-left connections (dashed red edges) are rarely active. By restricting our attention to the subspace of states in which left and right dynamics are independent we get a local causal model (\textbf{center right}) with two components that can be considered separately for training and inference.}
	\label{fig_local_factorization}
\end{figure}

\paragraph{LCMs}
Our key insight is that for each pair of nodes $(V^i_t, S^j_{t+1})$ with $V^i_t \in \parents(S^j_{t+1})$ in $\G$, there often exists a large subspace $\localSA \subset \S \times \A$ for which $S^j_{t+1} \indep 
V^i_t \given \parents(S^j_{t+1})\setminus V^i_t, (s_t, a_t) \in \localSA$.
For example, in case of a two-armed robot (Figure \ref{fig_local_factorization}), there is a large subspace of states in which the two arms are too far apart to influence each other physically. 
Thus, if we restrict our attention to $(s_t, a_t) \in \localSA$, we can consider a \textit{local} causal model $\M^{\localSA}_t$ whose local DAG $\G^{\localSA}$ is strictly sparser than the global DAG $\G$, as the structural minimality assumption applied to $\G^{\localSA}$ implies that there is no edge from $V^i_t$ to $S^j_{t+1}$. More generally, for any subspace $\L \subseteq S \times A$, we can induce the Local Causal Model (LCM) $\M^\L_t = \langle V^\L_t, U^\L_t, \F^\L \rangle$ with DAG $\G^\L$ from the global model $\M_t$ as:
\begin{itemize}[leftmargin=0.3in,labelsep=0.1in]
\item $V^\L_t = \{V^{\L,i}_{t[+1]}\}_{i=0}^{2n+m}$, where $P(V^{\L,i}_{t[+1]}) = P(V^i_{t[+1]} \given (s_t, a_t) \in \L)$.
\item $U^\L_t = \{U^{\L,i}_{t[+1]}\}_{i=0}^{2n+m}$, where $P(U^{\L,i}_{t[+1]}) = P(U^i_{t[+1]}\given (s_t, a_t) \in \L)$.
\item $\F^\L = \{f^{\L,i}\}_{i=0}^{2n+m}$, where $f^{\L,i} = f^i\restrict{\L}$ ($f^i$ with range of input variables restricted to $\L$). 
Due to structural minimality, the signature of $f^{\L,i}$ may shrink (as the range of the relevant variables is now restricted to $\L$), and corresponding edges in $\G$ will not be present in $\G^\L$.\footnote{As a trivial example, if $f^i$ is a function of binary variable $V^j$, and $\L = \{(s, a)\given V^j = 0\}$, then $f^{\L,i}$ is not a function of $V^j$ (which is now a constant), and there is no longer an edge from $V^j$ to $V^i$ in $\G^\L$.}
\end{itemize}

In case of the two-armed robot, conditioning on the arms being far apart simplifies the global DAG to a local DAG with two connected components (Figure \ref{fig_local_factorization}). This can make counterfactual reasoning considerably more efficient: given a factual situation in which the robot's arms are far apart, we can carry out separate counterfactual reasoning about each arm.

\paragraph{Leveraging LCMs} To see the efficiency therein, consider a general case with global causal model $\M$.
To answer the counterfactual question, ``\textit{what might the transition at time $t$ have looked like if component $S_t^i$ had value $x$ instead of value $y$?}'', we would ordinarily apply Pearl's do-calculus to $\M$ to obtain submodel $\M_{\DO(S_t^i = x)} = \langle V, U, \F_x \rangle$, where $\F_x = \F\setminus f^i \cup \{S_t^i = x\}$ and incoming edges to $S_t^i$ are removed from $\G_{\DO(S_t^i = x)}$ \cite{pearl2009causal}. The component distributions at time $t+1$ can be computed by reevaluating each function $f^j$ that depends on $S_t^i$. When $S_t^i$ has many children (as is often the case in the global $\G$), this requires one to estimate outcomes for many structural equations 
$\{f^j | V^j \in \text{Children}(V^i_t)\}$. 
But if both the original value of $S_t$ (with $S_t^i = y$) and its new value (with $S_t^i = x$) are in the set $\L$, the intervention is ``within the bounds'' of local model $\M^\L$ and we can instead work directly with local submodel $\M^\L_{\DO(S_t^i = x)}$ (defined accordingly). The validity of this follows from the definitions: since $f^{\L,j} = f^j\restrict{\L}$ for all of $S_t^i$'s children, the nodes $V_t^k$ for $k \not= i$ at time $t$ are held fixed, and the noise variables at time $t+1$ are unaffected, the distribution at time $t+1$ is the same under both models. When $S_t^i$ has fewer children in $\M^\L$ than in $\M$, this reduces the number of structural equations that need to be considered.

\section{Counterfactual Data Augmentation} \label{section_coda}

We hypothesize that local causal models will have several applications, and potentially lead to improved agent designs, algorithms, and interpretability. In this paper we focus on improving off-policy learning in RL by exploiting causal independence in local models for \textbf{Counterfactual Data Augmentation (CoDA)}. CoDA augments real data by making counterfactual modifications to a subset of the causal factors at time $t$, leaving the rest of the factors untouched. Following the logic outlined in the Subsection \ref{subsection_locally_factored}, this can understood as manufacturing ``fake'' data samples using the counterfactual model $\M^{[\L]}_{\DO(S_t^{i\mydots j} = x)}$, where we modify the causal factors $S_t^{i\mydots j}$ and resample their children. While this is always possible using a model-based approach if we have good models of the structural equations, it is particularly nice when the causal mechanisms are independent, as we can do counterfactual reasoning 
directly by reusing subsamples from observed trajectories. 

\begin{definition} 
The causal mechanisms represented by subgraphs $\G_i, \G_j \subset \G$ are \textbf{independent} when $\G_i$ and $\G_j$ are disconnected in $\G$.
\end{definition}

When $\G$ is divisible into two (or more) connected components, we can think of each subgraph as an independent causal mechanism that can be reasoned about separately. 

\begin{figure}[!t]
	\centering
	\includegraphics[width=\textwidth]{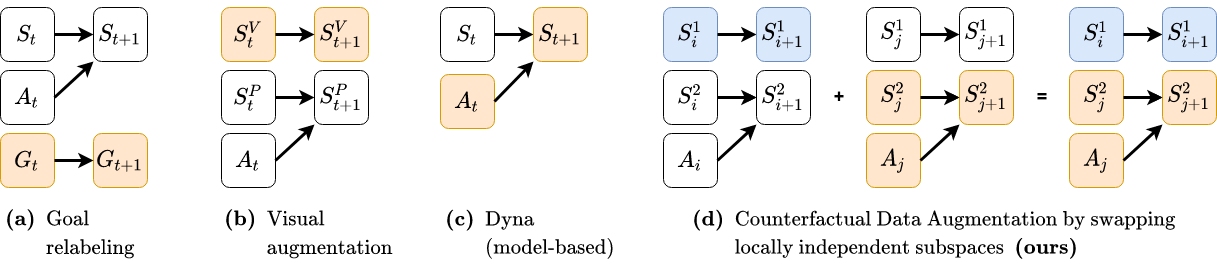}
	\vspace{-\baselineskip}
	\caption{\textit{Four instances of CoDA; orange nodes are relabeled, noise variables omitted for clarity.} \textbf{First:} Goal relabeling \cite{kaelbling1993learning}, including HER \cite{andrychowicz2017hindsight}, augments transitions with counterfactual goals. \textbf{Second:} Visual feature augmentation \cite{andrychowicz2020learning,laskin2020reinforcement} uses domain knowledge to change visual features $S_t^V$ (such as textures, lighting, and camera positions) that the designer knows do not impact the physical state $S^P_{t+1}$. \textbf{Third:} Dyna \cite{sutton1991dyna}, including MBPO \cite{janner2019trust}, augments real states with new actions and resamples the next state using a learned dynamics model. \textbf{Fourth (ours):} Given two transitions that share local causal structures, we propose to swap connected components to form new transitions.}
	\label{fig_coda_examples}
\end{figure}

Existing data augmentation techniques can be interpreted as specific instances of CoDA (Figure \ref{fig_coda_examples}). For example, goal relabeling \cite{kaelbling1993learning}, as used in Hindsight Experience Replay (HER) \cite{andrychowicz2017hindsight} and Q-learning for Reward Machines \cite{icarte2018using}, exploits the independence of the goal dynamics $G_t \mapsto G_{t+1}$ (identity map) and the next state dynamics $S_t \times A_{t} \mapsto S_{t+1}$ in order to relabel the goal variable $G_t$ with a counterfactual goal. While the goal relabeling is done model-free, we typically assume knowledge of the goal-based reward mechanism $G_t \times S_t \times A_t \times S_{t+1} \mapsto R_{t+1}$ to relabel the reward, ultimately mixing model-free and model-based reasoning. Similarly, visual feature augmentation, as used in reinforcement learning from pixels \cite{laskin2020reinforcement,kostrikov2020image} and sim-to-real transfer \cite{andrychowicz2020learning}, exploits the independence of the physical dynamics $S^P_t \times A_t \mapsto S^P_{t+1}$ and visual feature dynamics $S^V_t \mapsto S^V_{t+1}$ such as textures and camera position, assumed to be static ($S^V_{t+1} = S^V_t$), to counterfactually augment visual features. Both goal relabeling and visual data augmentation rely on \textit{global} independence relationships.

We propose a novel form of knowlen particular, we observe that whenever an environment transition is within the bounds of some local model $\M^\L$ whose graph $\G^\L$ has the locally independent causal mechanism $\G_i$ as a disconnected subgraph (note: $\G_i$ itself need not be connected), that transition contains an unbiased sample from $\G_i$. Thus, 
given two transitions in $\L$, 
we may mix and match the samples of $\G_i$ to generate counterfactual data, 
so long as the resulting transitions are themselves in $\L$. 

\begin{remark} How much data can we generate using our CoDA algorithm? If we have $n$ independent samples from subspace $\L$ whose graph $\G^\L$ has $m$ connected components, we have $n$ choices for each of the $m$ components, for a total of $n^m$ CoDA samples---an \textbf{exponential} increase in data! One might term this the ``blessing of independent subspaces.''
\end{remark}
\begin{remark} Our discussion has been at the level of a single transition (time slice $(t, t+1)$), which is consistent with the form of data that RL agents typically consume. But we could also use CoDA to mix and match locally independent components over several time steps (see, e.g., Figure \ref{fig_coda_intro_diagram}).
\end{remark}
\begin{remark} \label{remark_distributional_shift} As is typical, counterfactual reasoning changes the data distribution. While off-policy agents are typically robust to distributional shift, future work might explore different ways to control or prioritize the counterfactual data distribution \cite{schaul2015prioritized,kumar2020discor}. We note, however, that certain prioritization schemes may introduce selection bias \cite{hernan2020causal}, effectively entangling otherwise independent causal mechanisms (e.g., HER's ``future'' strategy \cite{andrychowicz2017hindsight} may introduce ``hindsight bias'' \cite{lanka2018archer,schroecker2020universal}).
\end{remark}
\begin{remark} The global independence relations relied upon by goal relabeling and image augmentation are incredibly general, as evidenced by their wide applicability. We posit that certain local independence relations are similarly general. For example, the physical independence of objects separated by space (the billiards balls of Figure \ref{fig_coda_intro_diagram}, the two-armed robot of Figure \ref{fig_local_factorization}, and the environments used in Section \ref{sec:experiments}), and the independence between an agent's actions and the truth of (but not belief about) certain facts the agent is ignorant of (e.g., an opponent's true beliefs).
\end{remark}

\paragraph{Implementing CoDA}

\newcommand{\Var}[1]{\texttt{#1}}
\begin{algorithm*}[t]\small
	\caption{Mask-based Counterfactual Data Augmentation (CoDA)} \label{alg_coda}
	\begin{small}\vspace{-0.15in}
	\begin{multicols}{2}\noindent
		\begin{algorithmic}
			\Function{Coda}{transition $\Var{t1}$, transition $\Var{t2}$}:
			\vspace{0.01in}
    		\State $\Var{s1}, \Var{a1}, \Var{s1'} \gets \Var{t1}$
	    	\State $\Var{s2}, \Var{a2}, \Var{s2'} \gets \Var{t2}$
	    	\State $\Var{m1}, \Var{m2} \gets \textsc{Mask}(\Var{s1}, \Var{a1}), \textsc{Mask}(\Var{s2}, \Var{a2})$
	    	\State $\Var{D1} \gets  \textsc{Components}(\Var{m1})$
 		    \State $\Var{D2} \gets  \textsc{Components}(\Var{m2})$
		    \State $\Var{d} \gets$ random sample from $(\Var{D1}$ $\cap$ $\Var{D2})$
		    \State $\Var{\~s}, \Var{\~a}, \Var{\~s'} \gets \textrm{copy}(\Var{s1}, \Var{a1}, \Var{s1'})$
		    \State $\Var{\~s[d]}, \Var{\~a[d]}, \Var{\~s'[d]} \gets \Var{s2[d]}, \Var{a2[d]}, \Var{s2'[d]}$
		    \State $\Var{\~D} \gets \textsc{Components}(\textsc{Mask}(\Var{\~s}, \Var{\~a}))$
		    \State \textbf{return} $(\Var{\~s},\Var{\~a},\Var{\~s'})$ \textbf{if} $\Var{d} \in \Var{\~D}$ \textbf{else} $\emptyset$
			\EndFunction
		\end{algorithmic}
		\columnbreak
		\begin{algorithmic}
		\Function{Mask}{state \Var{s}, action \Var{a}}:
			\begin{adjustwidth}{\algorithmicindent}{0pt}
			Returns $(n+m)\times(n)$ matrix indicating if the $n$ next state components (columns) locally depend on the $n$ state and $m$ action components (rows). 
			\end{adjustwidth}
		\EndFunction
		\Statex
		\vspace{-0.01in}
		\Function{Components}{mask \Var{m}}:
			\begin{adjustwidth}{\algorithmicindent}{0pt}
			Using the mask as the adjacency matrix for $\G^\L$ (with dummy columns for next action), finds the set of connected components $C = \{C_j\}$, and returns the set of independent components \newline $D = \{\G_i = \bigcup_k \mathcal{C}^i_{k}\given\mathcal{C}^i \subset \textrm{powerset}(C)\}$.
			\end{adjustwidth}
    	\EndFunction
		\end{algorithmic}
		\end{multicols}\vspace{-0.12in}
	\end{small}
\end{algorithm*}

We implement CoDA, as outlined above and visualized in Figure \ref{fig_coda_examples}(d), as a function of two factual transitions and a mask function $M(s_t, a_t): \S \times \A \to \{0, 1\}^{(n+m)\times n}$ that represents the adjacency matrix of the sparsest local causal graph $\G^\L$ such that $\L$ is a neighborhood of $(s_t, a_t)$.%
\footnote{If Jacobian $\partial P / \partial x$ exists at $x = (s_t, a_t)$, the ground truth $M(s_t, a_t)$ equals $|(\partial P / \partial x)^T| > 0$.} 
We apply $M$ to each transition to obtain local masks $\texttt{m}_1$ and $\texttt{m}_2$, compute their connected components, and 
swap independent components $\G_i$ and $\G_j$ (mutually disjoint and collectively exhaustive groups of connected components)
between the transitions to produce a counterfactual proposal. We then apply $M$ to the counterfactual $(\tilde s_t, \tilde a_t)$ to validate the proposal---if the counterfactual mask $\texttt{\~ m}$ shares the same graph partitions as $\texttt{m}_1$ and $\texttt{m}_2$, we accept the proposal as a CoDA sample. See Algorithm \ref{alg_coda}. 

Note that masks $\texttt{m}_1$, $\texttt{m}_2$ and $\texttt{\~ m}$ correspond to different neighborhoods $\L_1, \L_2$ and $\tilde\L$, so it is not clear that we are ``within the bounds'' of any model $\M^\L$ as was required in Subsection \ref{subsection_locally_factored} for valid counterfactual reasoning. 
To correct this discrepancy we use the following proposition and additionally require the causal mechanisms (subgraphs) for independent components $\G_i$ and $\G_j$ to share structural equations in each local neighborhood: $f^{\L_1,i} = f^{\L_2,i} = f^{\tilde\L,i}$ and $f^{\L_1,j} = f^{\L_2,j} = f^{\tilde\L,j}$.\footnote{To see why this is not trivially true, imagine there are two rooms, one of which is icy. In either room the ground conditions are locally independent of movement dynamics, but not so if we consider their union.}  This makes our reasoning valid in the local subspace $\L^* = \L_1 \cup \L_2 \cup \tilde\L$.
See Appendix \ref{appdx_proposition} for proof. 

\begin{proposition}\label{proposition_union_of_local_sets}
The causal mechanisms represented by $\G_i, \G_j \subset \G$ are independent in $\G^{\L_1 \cup \L_2}$ if and only if  $\G_i$ and $\G_j$ are independent in both $\G^{\L_1}$ and $\G^{\L_2}$, and $f^{\L_1,i} = f^{\L_2,i}, f^{\L_1,j} = f^{\L_2,j}$.
\end{proposition}

Since CoDA only modifies data within local subspaces, this biases the resulting replay buffer to have more factorized transitions.
In our experiments below, we specify the ratio of observed-to-counterfactual data heuristically to control this selection bias, but find that off-policy agents are reasonably robust to large proportions of CoDA-sampled trajectories.
We leave a full characterization of the selection bias in CoDA to future studies, noting that knowledge of graph topology was shown to be useful in mitigating selection bias for causal effect estimation \citep{bareinboim2012controlling, bareinboim2014recovering}.

\paragraph{Inferring local factorization} 
While the ground truth mask function $M$ may be available in rare cases as part of a simulator, the general case either requires a domain expert to specify an approximate causal model (as in goal relabeling and visual data augmentation) or requires the agent to learn the local factorization from data.
Given how common independence due to physical separation of objects is, the former option will often be available. 
In the latter case, we note that the same data could also be used to learn a forward model. Thus, there is an implicit assumption in the latter case that learning the local factorization is easier than modeling the dynamics. We think this assumption is rather mild, as an accurate forward dynamics model would subsume the factorization, and we provide some empirical evidence of its validity in Section \ref{sec:experiments}.

Learning the local factorization is similar to conditional causal structure discovery \cite{spirtes2000causation, runge2019inferring, peters2017elements}, conditioned on neighborhood
$\L$ of $(s_t, a_t)$, 
except that the same structural equations must be applied globally (if the structural equations were conditioned on 
$\L$, 
Proposition \ref{proposition_union_of_local_sets} would fail). 
As there are many algorithms for general structure discovery \cite{spirtes2000causation,runge2019inferring}, and the arrow of time simplifies the inquiry \cite{granger1969investigating,peters2017elements}, there may be many ways to approach this problem.
For now, we consider a generalization of the global network mask approach used by MADE \cite{germain2015made} (eq. 10) for autoregressive distribution modeling and GraN-DAG \cite{lachapelle2019gradient} (eq. 6) for causal discovery, which additionally conditions the mask on the current state and action. 

This approach computes a locally conditioned network mask $M(s_t, a_t)$ by taking the matrix product of locally conditioned layer masks: $M(s_t,a_t) = \Pi_{\ell=1}^LM_\ell(s_t, a_t)$. This mask can be understood as an upper bound on the network's absolute Jacobian (see Appendix \ref{sec:inferring_local_factorization}).
Again, there may be several models allow one to compute conditional layer masks. We tested two such models: a mixture of MLP experts and a single-head set transformer architecture \cite{vaswani2017attention,lee2018set}. Each is described in more detail in Appendix \ref{sec:inferring_local_factorization}. Both are \emph{trained} to model forward dynamics using an L2 prediction loss and induce a sparse network mask either via a sparsity penality (in case of the mixture of experts model) or via a sparse attention mechanism (in case of the set transformer). 
In preliminary experiments (Appendix \ref{sec:inferring_local_factorization}) we found that the set transformer performed better, and proceed to use it in our main experiments (Section \ref{sec:experiments}). The set transformer uses the attention mask at each layer as the layer mask for that layer, so that the network mask is simply the product of the attention masks. 
Though trained to model forward dynamics, the CoDA models are used by the agent to \emph{infer} local factorization rather than to directly sample future states as is typical in model-based RL.
We found this produced reasonable results in the tested domains (below).
See Appendix \ref{sec:inferring_local_factorization} for details. Future work should consider other approaches to inferring local structure such as graph neural networks \cite{kipf2019contrastive,burgess2019monet}.

\section{Experiments}\label{sec:experiments}

Our experiments evaluate CoDA in the online, batch, and goal-conditioned settings, in each case finding that CoDA significantly improves agent performance as compared to non-CoDA baselines. Since CoDA only modifies an agent's training data, we expect these improvements to extend to other off-policy task settings in which the state space can be accurately disentangled. Below we outline our experimental design and results, deferring specific details and additional results to Appendix \ref{sec:training_details}.

\begin{figure}[!b]
	\centering
	\includegraphics[width=\textwidth]{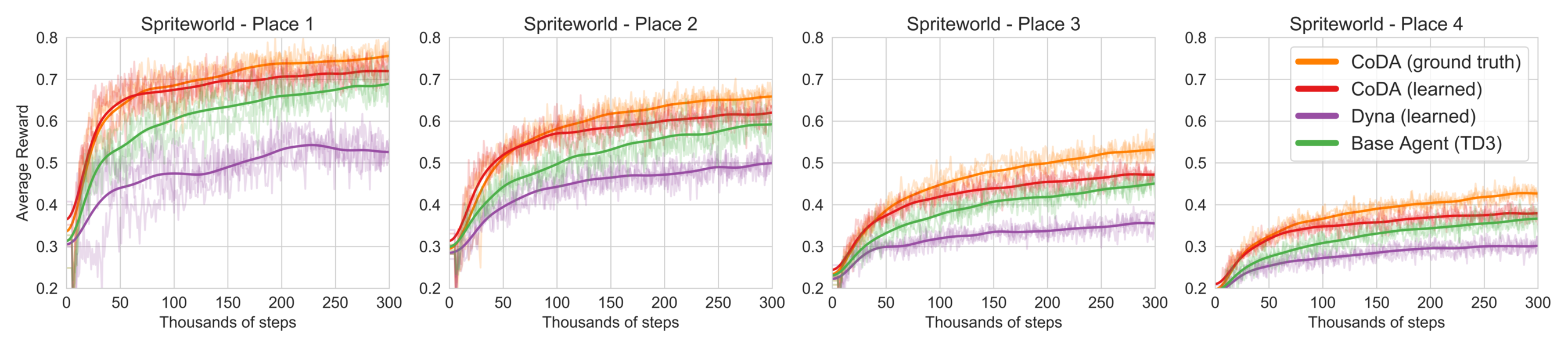}
	\caption{\textbf{Standard online RL} (3 seeds): CoDA with the ground truth mask always performs the best, validating our basic idea. CoDA with a pretrained model also offers a significant early boost in sample efficiency and maintains its lead over the base TD3 agent throughout training. Using the same model to generate data directly (a la Dyna \cite{sutton1991dyna}) performs poorly, suggesting significant model bias.}
	\label{fig_rl_results_plot}
\end{figure}

\paragraph{Standard online RL}

\begin{wrapfigure}{R}{0.17\textwidth}
    \vspace{-\baselineskip}
	\centering
    \captionsetup{width=0.16\textwidth}
	\includegraphics[width=0.16\textwidth,height=0.15\textwidth]{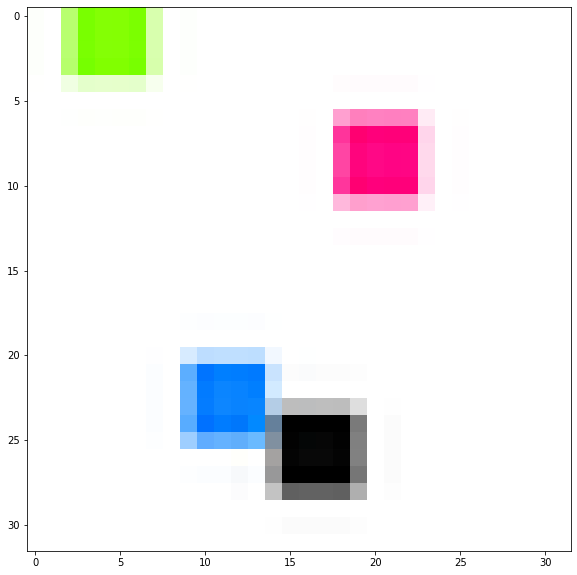}
	\label{fig_env_spriteworld}
    \vspace{-\baselineskip}
\end{wrapfigure}

We extend \texttt{Spriteworld} \cite{watters2019cobra} to construct a ``bouncing ball'' environment (right), that consists of multiple objects (sprites) that move and collide within a confined 2D canvas. 
We use tasks of varying difficulty, where the agent must navigate $N \in \{1, 2, 3, 4\}$ of 4 sprites to their fixed target positions. The agent receives reward of $1/N$ for each of the $N$ sprites placed; e.g., the hardest task (Place 4) gives $1/4$ reward for each of 4 sprites placed.  
For each task, we use CoDA to expand the replay buffer of a TD3 agent \cite{fujimoto2018addressing} by about 8 times. We compare CoDA with a ground truth masking function (available via the \texttt{Spriteworld} environment) and learned masking function to the base TD3 agent, as well as a Dyna agent that generates additional training data by sampling from a model. 
For fair comparison, we use the same transformer used for CoDA masks for Dyna, which we pretrain using approximately 42,000 samples from a random policy. 
As in HER, we assume access to the ground truth reward function to relabel the rewards. The results in Figure \ref{fig_rl_results_plot} show that both variants of CoDA significantly improve sample complexity over the baseline. By contrast, the Dyna agent suffers from model bias, even though it uses the same model as CoDA.

{\small
\begin{table*}[t]
\newcommand{\smol}[1]{{\scriptsize\texttt{#1}}}
\centering\small
\begin{tabular}{c@{\hskip 0.2in}cc@{\hskip 0.16in}|@{\hskip 0.16in}cccc}
	\toprule
$|\mathcal{D}|$	&Real data& MBPO&
\multicolumn{4}{c}{Ratio of Real:CoDA [:MBPO] data (ours)}\\
$(1000s)$ &   1\smol{R}  &   1\smol{R}:1\smol{M} &  1\smol{R}:1\smol{C}  &  1\smol{R}:3\smol{C}  &  1\smol{R}:5\smol{C} & 1\smol{R}:3\smol{C}:1\smol{M} \\
	\midrule
$\hphantom{0}25$ & $13.2 \pm 0.7$  & $18.5 \pm 1.5$  & $43.8 \pm 2.8$  & $40.9 \pm 2.5$  & $38.4 \pm 4.9$  & $\mybm{46.8} \pm 3.1$  \\
$\hphantom{0}50$ & $22.8 \pm 3.0$  & $36.6 \pm 4.3$  & $66.6 \pm 3.8$  & $64.4 \pm 3.1$  & $62.5 \pm 3.5$  & $\mybm{70.4} \pm 3.8$  \\
$\hphantom{0}75$ & $43.2 \pm 4.9$  & $46.0 \pm 4.7$  & $73.4 \pm 2.8$  & $\mybm{76.7} \pm 2.6$  & $75.0 \pm 3.4$  & $74.6 \pm 3.2$  \\
$100$ & $63.0 \pm 3.1$  & $66.4 \pm 4.9$  & $77.8 \pm 2.0$  & $\mybm{82.7} \pm 1.5$  & $76.6 \pm 3.0$  & $73.7 \pm 2.9$  \\
$150$ & $77.4 \pm 1.2$  & $72.6 \pm 5.6$  & $82.2 \pm 1.8$  & $\mybm{85.8} \pm 1.4$  & $84.2 \pm 1.0$  & $79.7 \pm 3.6$  \\
$250$ & $78.2 \pm 2.7$  & $77.9 \pm 2.4$  & $85.0 \pm 2.9$  & $\mybm{87.8} \pm 1.8$  & $87.0 \pm 1.0$  & $78.3 \pm 4.9$  \\
	\bottomrule
	\vspace{-0.12in}
\end{tabular}
\caption{\textbf{Batch RL} (10 seeds): Mean success ($\pm$ standard error, estimated using 1000 bootstrap resamples) on \texttt{Pong} environment. CoDA with learned masking function more than doubles the effective data size, resulting in a 3x performance boost at smaller data sizes. Note that a 1\smol{R}:5\smol{C} Real:CoDA ratio performs slightly worse than a 1\smol{R}:3\smol{C} ratio due to distributional shift (Remark \ref{remark_distributional_shift}).
}\label{fig_batch_rl_and_fetch}
\vspace{-\baselineskip}
\end{table*}
}

\paragraph{Batch RL}

\begin{wrapfigure}{R}{0.17\textwidth}
	\centering
    \captionsetup{width=0.16\textwidth}
	\includegraphics[width=0.16\textwidth,height=0.15\textwidth]{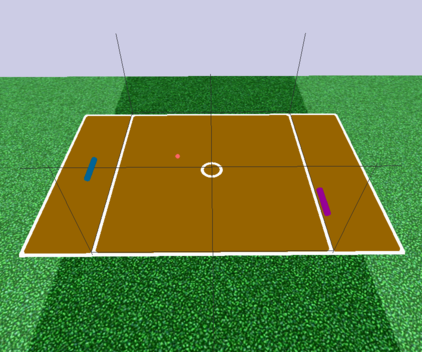}
	\label{fig_env_pong}
    \vspace{-\baselineskip}
\end{wrapfigure}

A natural setting for CoDA is batch-constrained RL, where an agent has access to an existing transition-level dataset, but cannot collect more data via exploration \cite{fujimoto2018off,levine2020offline}.  Another reason why CoDA is attractive in this setting is that there is no \textit{a priori} reason to prefer the given batch data distribution to a counterfactual one.  
For this experiment we use a continuous control \texttt{Pong} environment based on \texttt{RoboschoolPong} \cite{klimov2017roboschool}. The agent must hit the ball past the opponent, receiving reward of +1 when the ball is behind the opponent's paddle, -1 when the ball is behind the agent's paddle, and 0 otherwise. 
Since our transformer model performed poorly when used as a dynamics model, our Dyna baseline for batch RL adopts a state-of-the-art architecture \cite{janner2019trust} that employs a 7-model ensemble (MBPO).
We collect datasets of up to 250,000 samples from an pretrained policy with added noise. For each dataset, we train both mask and reward functions (and in case of MBPO, the dynamics model) on the provided data and use them to generate different amounts of counterfactual data. We also consider combining CoDA with MBPO, by first expanding the dataset with MBPO and then applying CoDA to the result. We train the same TD3 agent on the expanded datasets in batch mode for 500,000 optimization steps. The results in Table \ref{fig_batch_rl_and_fetch} show that with only 3 state factors (two paddles and ball), applying CoDA is approximately equivalent to doubling the amount of real data.

\begin{figure*}[!h]
    \centering
    \begin{minipage}{0.24\textwidth}
    \centering
    \includegraphics[width=\textwidth]{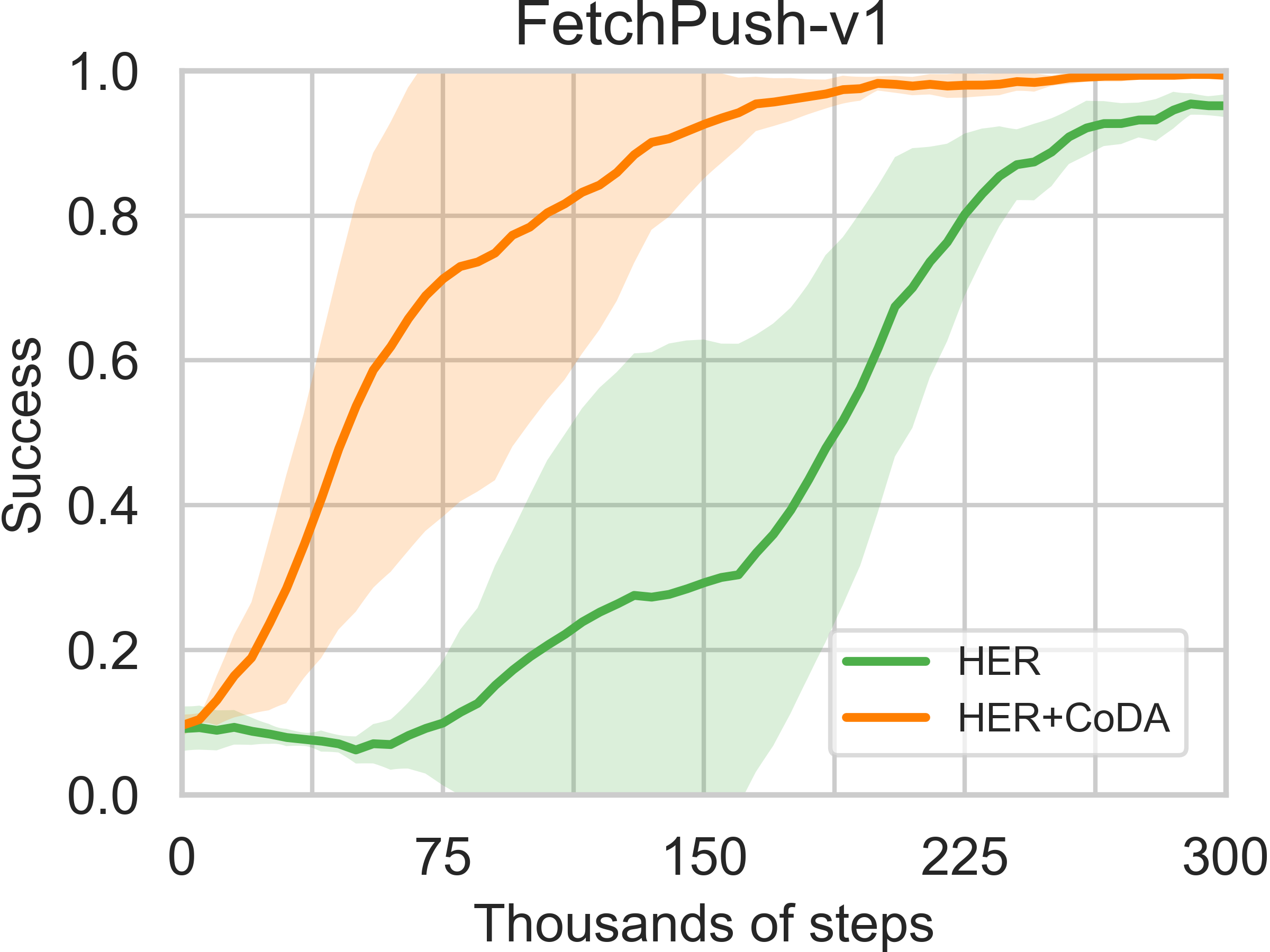}
    \end{minipage}%
    \begin{minipage}{0.01\textwidth}
   	\hphantom{X}
    \end{minipage}%
    \begin{minipage}{0.24\textwidth}
    \centering
    \includegraphics[width=\textwidth]{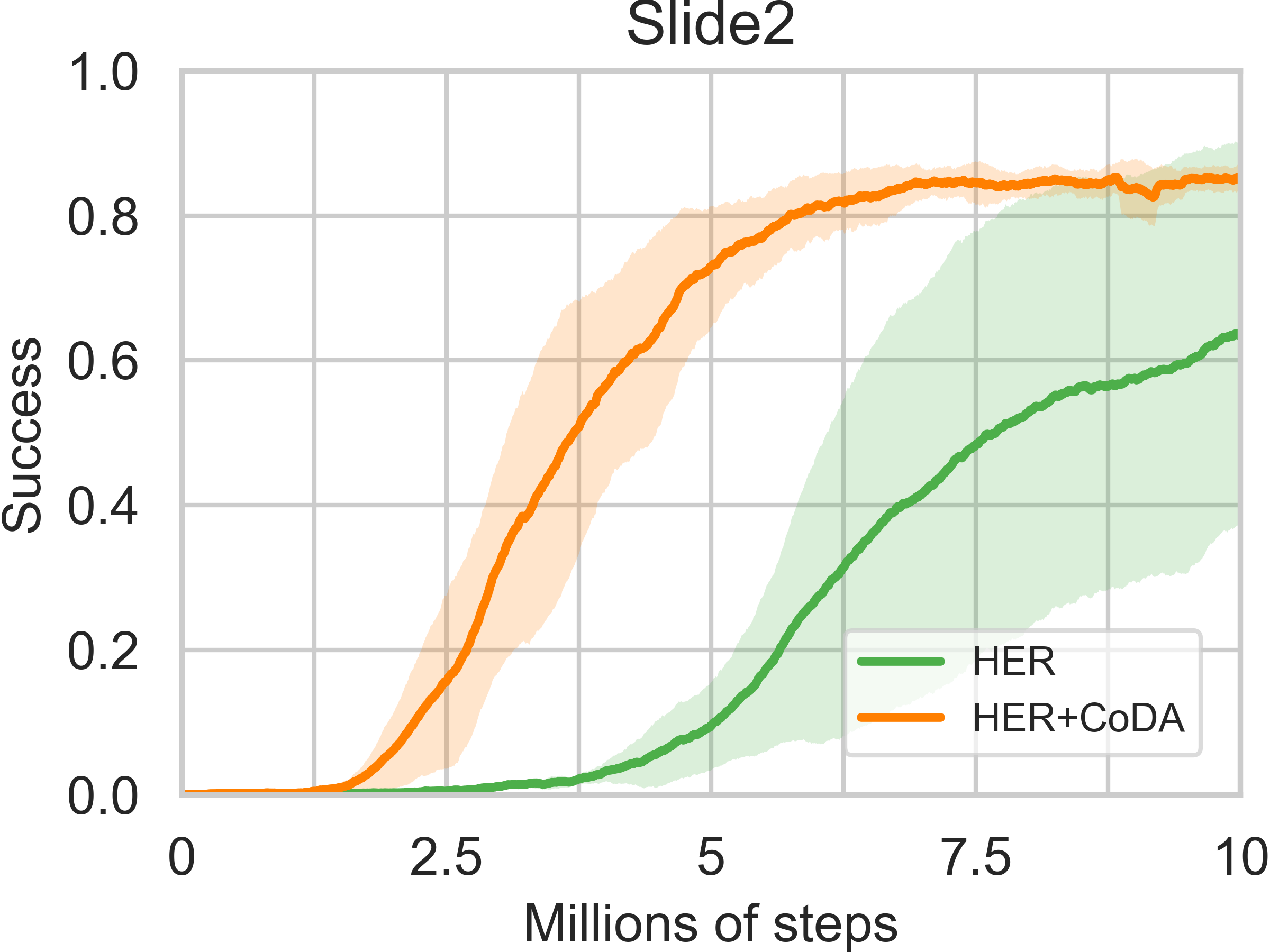}
    \end{minipage}%
    \begin{minipage}{0.04\textwidth}
	\hphantom{X}
    \end{minipage}%
    \begin{minipage}{0.47\textwidth}
    \vspace{0.08in}
    
    \renewcommand{\figurename}{Fig.}
    
    \caption{\textbf{Goal-conditioned RL} (5 seeds): In \texttt{FetchPush} and the challenging \texttt{Slide2} environment, a HER agent whose dataset has been enlarged with CoDA approximately doubles the sample efficiency of the base HER agent.}\label{fig_goal_conditioned_results}
    \end{minipage}
    \vspace{-\baselineskip}
\end{figure*}

\begin{wrapfigure}{r}{0.17\textwidth}
    \vspace{-\baselineskip}
	\centering
    \captionsetup{width=0.16\textwidth}
	\includegraphics[width=0.16\textwidth,height=0.15\textwidth]{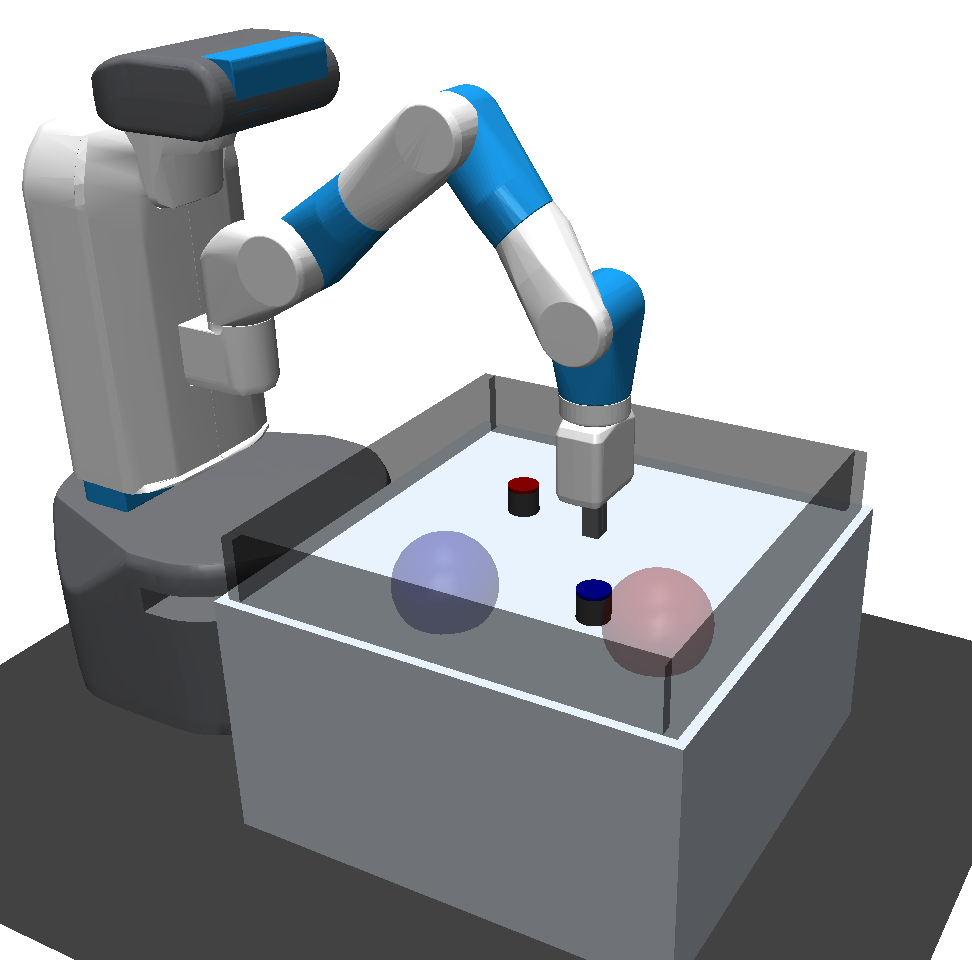}
	\label{fig_env_fetch}
    \vspace{-\baselineskip}
\end{wrapfigure}
\paragraph{Goal-conditioned RL} As HER \cite{andrychowicz2017hindsight} is an instance of prioritized CoDA that greatly improves sample efficiency in sparse-reward tasks, can our unprioritized CoDA algorithm further improve HER agents? We use HER to relabel goals on real data only, relying on random CoDA-style goal relabeling for CoDA data.
After finding that CoDA obtains state-of-the-art results in \texttt{FetchPush-v1} \cite{plappert2018multi}, we show that CoDA also accelerates learning in a novel and significantly more challenging \texttt{Slide2} environment, where the agent must slide two pucks onto their targets (Figure \ref{fig_goal_conditioned_results}). For this experiment, we specified a heuristic mask using domain knowledge (``objects are disentangled if more than 10cm apart'') that worked in both \texttt{FetchPush} and \texttt{Slide2} despite different dynamics.

\section{Related Work}\label{section_related_work}

Factored MDPs \cite{guestrin2003efficient,hallak2015off,weber2019credit} consider MDPs where state variables are only influenced by a fixed subset of ``parent'' variables at the previous timestep.
The notion of ``context specific independence'' (CSI), which was used to compactly represent single factors of a Bayes net \citep{boutilier2013context} or MDP \citep{boutilier1995exploiting} for efficient inference and model storage,\footnote{Note that these methods were proposed to efficiently encode conditional probability tables at the graph nodes, requiring that all variables considered be discrete; CoDA on the other hand works in continuous tasks.} is closely related to the local factorizations we study in this paper.
CSI can be understood as going one step beyond CoDA, exploiting not only knowledge of the local factorization, but also the structural equations at play in the local factorization; CSI could be leveraged for model-based RL approaches where faithful models of factored dynamics can be realized.
Object-oriented and relational approaches to RL and prediction \cite{diuk2008object,goyal2019recurrent,kipf2019contrastive,li2019towards,zambaldi2018relational,zhu2018object} represent the dynamics as a set of interacting entities. Factored actions and policies have been used to formulate dimension-wise policy gradient baselines in standard and multi-agent settings \cite{foerster2018counterfactual, lowe2017multi,wu2018variance}. 

A growing body of work applies causal reasoning the RL setting to improve sample efficiency, interpretability and learn better representations \cite{lu2018deconfounding,madumal2019explainable,rezende2020causally}. Particularly relevant is the work by \citet{buesing2018woulda}, which improves sample efficiency by using a causal model to sample counterfactual trajectories, thereby reducing variance of off-policy gradient estimates in a guided policy search framework. 
These counterfactuals use coarse-grained representations at the trajectory level, while our approach uses factored representations within a single transition.
Batch RL \cite{levine2020offline,fujimoto2018off,mandel2014offline} and  more generally off-policy RL \cite{watkins1992q,munos2016safe} are counterfactual by nature, and are particularly important when it is costly or dangerous to obtain on-policy data \cite{thomas2016data}. The use of counterfactual goals to accelerate learning goal-conditioned RL \cite{kaelbling1993learning,schaul2015universal,plappert2018multi} is what inspired our local CoDA algorithm.

Data augmentation is also widely used in supervised learning, and is considered a required best practice in high dimensional problems \citep{krizhevsky2012imagenet,laskin2020reinforcement,perez2017effectiveness}.
Heuristics for data augmentation often encode a causal invariance statement with respect to certain perturbations on the inputs.
Thus model performance on counterfactual/augmented data can be seen as a measure of \emph{robustness}. 
Assuring that models perform robustly in this sense is relevant 
to applications where \emph{fairness} is a concern, as counterfactuals can be used to achieve robust performance and debias data
\citep{garg2019counterfactual,park2018reducing,bolukbasi2016man}.

\section{Conclusion}
In this paper we proposed a local causal model (LCM) framework that captures the benefits of decomposition in settings where the global causal model is densely connected. We used our framework to design a local Counterfactual Data Augmentation (CoDA) algorithm that expands available training data with counterfactual samples by stitching together locally independent subsamples from the environment. Empirically, we showed that CoDA can more than double the sample efficiency and final performance of reinforcement learning agents in locally factored environments. 

There are several interesting avenues for future work. 
First, the sizable gap between ground truth and learned CoDA in our Spriteworld results suggest there is room for improvement in our approach to learning the masking function. Second, we have applied CoDA in a random, unprioritized fashion, but past work \cite{andrychowicz2017hindsight,schaul2015prioritized} suggests there is significant benefit to prioritization. Third, we have applied CoDA in a way that might be considered model-free, insofar as we reuse subsamples from the environment dynamics rather than generating samples using our models of the causal mechanims. However, our LCM formalism allows for mixing model-free and model-based reasoning, which could further improve sample efficiency. Fourth, we used fully-observable, disentangled state spaces with a fixed top-level decomposition, but ultimately we would like to deploy CoDA in partially observable, entangled settings (e.g. RL from pixels) with multiple dynamic, multi-level decompositions \cite{higgins2018scan,esmaeili2019structured,zaheer2017deep}. Unsupervised learning of factorized latent representations is an active area of research \cite{dundar2020unsupervised,kipf2019contrastive,locatello2018challenging,locatello2020weakly,watters2019cobra}, and it would be interesting to combine these methods with CoDA. 
Finally, it would be interesting to explore applications of our LCM framework to other areas such as interpretability \cite{lyu2019sdrl,mott2019towards}, exploration \cite{nair2018visual,wang2019monkeys}, and off-policy evaluation \cite{thomas2016data}.

\begin{ack}

We thank Jimmy Ba, Harris Chan, Seyed Kamyar Seyed Ghasemipour, James Lucas, David Madras, Yuhuai Wu and Lunjun Zhang for helpful comments and discussions. 
We also thank the anonymous reviewers for their feedback, which improved the final manuscript.
SP is supported by an NSERC PGS-D award. EC is a student researcher at Google Brain in Toronto.
Resources used in preparing this research were provided, in part, by the Province of Ontario, the Government of Canada through {CIFAR}, and companies sponsoring the Vector Institute ({\url{https://vectorinstitute.ai/partners}}).
\end{ack}

\section*{Broader Impact}

\paragraph{Considerations related to counterfactual reasoning}

CoDA transforms the observational data distribution into a counterfactual one. This incurs several risks and benefits, listed below. 
\begin{itemize}[leftmargin=0.3in,labelsep=0.1in]
\item Modern machine learning models and reinforcement learning agents often generalize poorly under distributional shift (sometimes called ``covariate shift'') \cite{li2018learning,cobbe2018quantifying}. CoDA has the potential to produce out-of-distribution data, that would never show up in the observational distribution. Thus, care should be taken when applying agent modules that have been trained only on observational data to counterfactual data, as their performance could decline sharply, thereby creating safety risks. We anticipate that work on uncertainty will be essential to controlling the risks associated with distributional shift \cite{snoek2019can}.
\item The fact that CoDA creates distributional shift can also provide benefits in the form of distributional robustness and fairness. Training on out-of-distribution data can make models more robust \cite{sinha2017certifying}, increasing their trustworthiness and practical applicability. As noted in Section \ref{section_related_work}, counterfactual reasoning can be leveraged in areas where \emph{fairness} is a concern. For example, \citep{park2018reducing} propose to reduce gender bias in natural language processing by generating counterfactual sentences with swapped gender pronouns. We anticipate that a version of CoDA that prioritizes fairness concerns in a similar manner could be applied in the reinforcement learning and computer vision contexts. 
\item In reinforcement learning specifically, a shift in data distribution requires either (1) the use of an off-policy algorithm, or (2) high variance off-policy corrections for on-policy algorithms. In the former case, it should be noted that in case of function approximation, even off-policy algorithms may be negatively affected by large shifts in their training distribution \cite{sutton2018reinforcement,fu2019diagnosing}. More work is needed to quantify these effects and their implications for agent performance. 
\end{itemize}

\paragraph{Improving RL in batch-constrained settings}
In many settings, such as medicine and education, obtaining large quantities of observational data using a random policy is prohibitively expensive and/or unethical \cite{thomas2016data,munos2016safe}.
As such, agents that can efficiently learn effective policies from batch-constrained data are needed, as are accurate ways to estimate agent performance from off-policy data \cite{mandel2014offline,oberst2019counterfactual,raghu2018behaviour}.
We see CoDA as complementary to both goals, as subspace swapping is a powerful tool to produce large quantities of counterfactual data given a modest observational dataset.
However, subspace swapping alone may be insufficient to generate plausible ``exploratory'' data for evaluating and learning new policies.
For example, medical records from certain demographic groups may be unavailable or improperly collected/labeled.
It is conceivable that a CoDA with a suitable prioritization scheme could compensate for such sample bias, but applied work in batch-constrained domains that characterizes the effect of sample bias on CoDA should nevertheless be carried out.

\paragraph{General considerations related to artificial agency} 

To the extent that CoDA is a general technique for improving the ability of artificial agents to achieve their goals, it inherits the potential risks and benefits associated with empowered artificial agency, including but not limited to: 
\begin{enumerate}[label=(\alph*),leftmargin=0.3in,labelsep=0.1in]
\item the pursuit of misguided or dangerous goals, whether due to mispecification by a benevolent principal, the self-serving motives of its principals, or interference by malicious parties or other deviations from proper intents,
\item the unsafe and improper pursuit of goals due to poor modeling or representation, resource constraints and lack of capacity, constraint mispecification, partial observability, or inadequate encoding and understanding of human values, and
\item improvements to capital processes and automation of human labor, which could improve economic efficiency and raise the overall social welfare, but also run the risk of increased inequality, workforce displacement, and technological unemployment. 
\end{enumerate}
The risk associated with points (a) and (b) may be exacerbated in case of CoDA due to the risks associated with counterfactual reasoning outlined above: to the extent that CoDA is done with a poorly fit local factorization model, or with a local factorization model that does not generalize well to the counterfactual distribution, this could cause the agent to pursue poorly formulated counterfactual (imagined) goals, or create causally invalid data that hurts agent performance.

{\small
\bibliographystyle{plainnat}
\bibliography{refs}
}

\newpage

\appendix

\section{Proof of Proposition 1}\label{appdx_proposition}

\begin{lemma}
If $V^j \in \parents^\L(V^i)$ in DAG $G^\L$ of (local) causal model $\M^\L$, and $\L \subset \mathcal{X}$, then $V^j \in \parents^\mathcal{X}(V^i)$ in DAG $G^\mathcal{X}$ corresponding to causal model $\M^\mathcal{X}$. 
\end{lemma}
\vspace{-\baselineskip}
\begin{proof}
By minimality, there exist $\{u^i, v^{-j}, v^j_1\}$ and $\{u^i, v^{-j}, v^j_2 \}$ with $v^{-j} \in \parents^\L(V^i)\setminus V^j$ for which $f^i(\{u^i, v^{-j}, v^j_1\}) \not= f^i(\{u^i, v^{-j}, v^j_2 \})$. Expand $\{v^{-j}, v^j_1\}$ and $\{v^{-j}, v^j_2\}$ to $(s_1, a_1), (s_2, a_2) \in \L$ (with any values of other components). But $\L \subset \mathcal{X}$, so $(s_1, a_1), (s_2, a_2) \in \mathcal{X}$ and it follows from minimality in $\mathcal{X}$ that $V^j \in \parents^\mathcal{X}(V^i)$.
\end{proof}

\begin{corollary}\label{corollary_sparser}
If $\L \subset \mathcal{X}$, $G^\L$ is sparser (has fewer edges) than $G^\mathcal{X}$. 
\end{corollary}\vspace{\baselineskip}

\begin{appdxProp}{1}
The causal mechanisms represented by $\G_i, \G_j \subset \G$ are independent in $\G^{\L_1 \cup \L_2}$ if and only if  $\G_i$ and $\G_j$ are independent in both $\G^{\L_1}$ and $\G^{\L_2}$, and $f^{\L_1,i} = f^{\L_2,i}, f^{\L_1,j} = f^{\L_2,j}$.
\end{appdxProp}
\vspace{-\baselineskip}
\begin{proof}

($\Rightarrow$)
If $\G_i$ and $\G_j$ are independent in $G^{\L_1 \cup \L_2}$, independence in $\G^{\L_1}$ and $\G^{\L_2}$ follows from Corollary \ref{corollary_sparser}. That $f^{\L_1,i} = f^{\L_2,i}$ (and $f^{\L_1,j} = f^{\L_2,j}$), on their shared domain, follows since each is a restriction of the same function $f^{\L_1 \cup \L_2,i}$ (or $f^{\L_1 \cup \L_2,j}$).

($\Leftarrow$)
Suppose $\G_i$ and $\G_j$ are independent in $\G^{\L_1}$ and $\G^{\L_2}$ but not $G^{\L_1 \cup \L_2}$. By the definition of independence applied to $G^{\L_1 \cup \L_2}$, we have that, without loss of generality, there is a $V_i \in \G_i, V_j \in \G_j$ with $V_j \in \parents^{\L_1 \cup \L_2}(V_i)$. Then, from the definition of minimality, it follows that there exist $(s_1, a_1), (s_2, a_2) \in \L_1 \cup \L_2$ that differ only in the value of $V_j$, and $u_i \in \textrm{range}(U_i)$ for which $f^i(s_1, a_1, u_i) \not= f^i(s_2, a_2, u_i)$. 

Clearly, if $(s_1, a_1)$ and $(s_2, a_2)$ are both in $\L_1$ (or $\L_2$), there will be an edge from $V_j$ to $V_i$ in $\G^{\L_1}$ (or $\G^{\L_2}$) and the claim follows by contradiction. Thus, the only interesting case is when, without loss of generality, $(s_1, a_1) \in \L_1$ and $(s_2, a_2) \in \L_2$. 
The key observation is that $(s_1, a_1)$ and $(s_2, a_2)$ differ only in the value of node $V_j \not\in \G_i$: since $\G_i$ is an independent causal mechanism in both $\G^{\L_1}$ and $\G^{\L_2}$ and the parents of $V_i$ take on the same values in each, we have that $f^i(s_1, a_1, u_i) = f^{i,\L_1}(s_1, a_1, u_i) = f^{i,\L_2}(s_2, a_2, u_i) = f^i(s_2, a_2, u_i)$ and the claim follows by contradiction. 
\end{proof}

\section{Additional Experiment Details}\label{sec:training_details}
This section provides training details for the experiments discussed in Section \ref{sec:experiments} as well as some additional results. Code is available at \url{https://github.com/spitis/mrl} \cite{mrl}.

\subsection{Online RL}
Here we detail the procedure used in the Online RL experiments in the Spriteworld environment.

\paragraph{Implementation} We work
with the original TD3 codebase, architecture, and hyperparameters (except batch size; see below), and focus our
efforts solely on modifying the agent’s training distribution.

\paragraph{Environment} We extend the base Spriteworld framework \cite{watters2019cobra} with (1) basic collisions (to induce a local factorization / so that a global factorization will not work), (2) a new continuous action space (2-dimensional), (3) a disentangled state renderer that returns the position and velocity of each sprite (a total of 16-dimensions in tasks with four sprites), (4) a mask renderer that returns the ground truth masking function (allows us to evaluate our masking function in Appendix \ref{sec:inferring_local_factorization}), and (5) a suite of partial and sparse reward tasks that we use for experiments. These extensions will included with the release of our code.

\paragraph{Data augmentation}
Every 1000 environment steps, we sample 2000 pairs of random transitions from the agent's replay buffer, and apply CoDA to produce a maximum of 5 unique CoDA samples per pair. We apply two forms of CoDA, using (1) an oracle / ground truth mask function that we back out of the simulator, and (2) the mask of a pre-trained transformer model (see Section \ref{sec:inferring_local_factorization}).
The mask function was trained using approximately 42,000 samples from a random policy (5/6 of 50,000, with the rest of the data used for validation). CoDA samples are added to a second CoDA replay buffer. For purposes of this experiment both buffers are have effectively infinite capacity (they are never filled). During training, the agent's batches are sampled proportionally from the real and CoDA replay buffers (this means that approximately 7/8 of the data that the agent trains on is counterfactual).

\paragraph{Baselines} 
In addition to the base TD3 agent and CoDA, we also use the transformer model that is used as a mask function to generate data by performing forward rollouts with a random policy, as in Dyna \cite{sutton1991dyna}. So that this baseline produces approximately the same number of samples as CoDA with the learned mask, we roll the model out for 5 steps from 1500 random samples from the replay buffer, again every 1000 environment steps. This produces 7500 model-based samples for every 1000 environment transitions.

Use of the transformer mask function requires setting the threshold value $\tau$, which we do by monitoring accuracy and F1 scores for sparsity prediction (as discussed in Appendix \ref{sec:inferring_local_factorization}) on validation data,
ultimately using the value $\tau = 0.05$.

\paragraph{Batch Size}
Since CoDA samples are plentiful we increase the agent's batch size from 256 to 1000 to allow it to train on more environment samples in the same number of training steps. We found that this slightly improved the performance of the base TD3 agent. An increase in batch size also allows the agent to see more of its own on-policy data in the face of many off-policy CoDA samples. 

\paragraph{Additional results in sparse reward task variants}

\begin{figure}[!t]
	\centering
	\includegraphics[width=\textwidth]{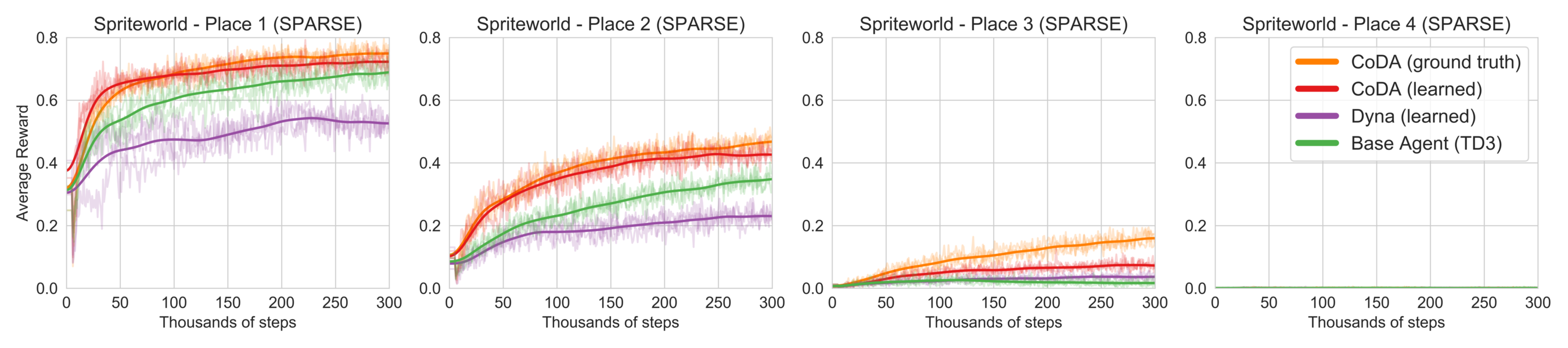}
	\caption{\textbf{Average reward on Sparse reward bouncing ball environment.} As in the partial reward case (Figure \ref{fig_rl_results_plot}), we observe that CoDA agents outperform the other agents (except in Place 4, where no agent achieves any reward).}
	\label{fig_rl_results_plot_sparse}
\end{figure}

In addition to the partial reward tasks described in Section \ref{sec:experiments}, we also tested CoDA in four sparse reward tasks of varying difficulty. These are the same as the partial reward tasks, except that a sparse reward of 1 is granted only when \textit{all} N sprites are in their target locations. While these tasks were much harder (and perhaps impossible in the case of Place 4 due to moving sprites), as shown in Figure \ref{fig_rl_results_plot_sparse}, the CoDA agents maintain a clear advantage.

\paragraph{Results plot} The results plot shows the 3 seeds in reduced opacity together with their smoothed mean.

\subsection{Batch RL}
Here we detail the procedure used in the Batch RL experiments in the \texttt{Pong} environment.

\paragraph{Implementation} This experiment works with a different codebase than the Spriteworld experiment, in order to simplify the use of CoDA in a Batch RL setting. Our experiment first builds the agent's dataset (consisting of real data, dyna data and/or CoDA data), then instantiates a TD3 agent by filling its replay buffer with the dataset. The replay buffer is always expanded to include the entire enlarged dataset (for the 5x CoDA ratio at 250,000 data size this means the buffer has 1.5E6 experiences). The agent is run for 500,000 optimization steps.

\paragraph{Hyperparameters} We used similar hyperparameters to the original TD3 codebase, with the following differences:
\begin{itemize}
\item We use a discount factor of $\gamma = 0.98$ instead of 0.99. 
\item Since Pong is a sparse reward task with $\gamma = 0.98$, we clip critic targets to $(-50, 50)$.
\item We use networks of size $(128, 128)$ instead of $(256, 256)$. 
\item As for our Spriteworld experiment, we use a batch size of $1000$. 
\end{itemize}

\paragraph{Environment} We base our \texttt{Pong} environment on \texttt{RoboSchoolPong-v1} \cite{klimov2017roboschool}. The original environment allowed the ball to teleport back to the center after one of the players scored, offered a small dense reward signal for hitting the ball, and included a stray ``timeout'' feature in the agent's state representation. We fix the environment so that the ball does not teleport, and instead have the episode reset every 150 steps, and also 10 steps after either player scores. The environment is treated as continuous and never returns a done signal that is not also accompanied by a \texttt{TimeLimit.truncated} indicator \cite{brockman2016openai}. We change the reward to be strictly sparse, with reward of $\pm 1$ given when the ball is behind one of the players' paddles. Finally, we drop the stray ``timeout'' feature, so that the state space is 12-dimensional, where each set of 4 dimensions is the x-position, y-position, x-velocity, and y-velocity of the corresponding object (2 paddles and one ball). 

\paragraph{Training the CoDA model}
Without access to a ground truth mask, we needed to train a masking function \ref{sec:inferring_local_factorization} to identify local disentanglement. We also forewent the ground truth reward, instead training our own reward classifier. In each case we used the batch dataset given, and so we trained different models for each random seed. For our masking model, we stacked two single-head transformer blocks (without positional encodings) and used the product of their attention masks as the mask. 
Each block consists of query $Q$, key $K$, and value $V$ networks that each have 3-layers of 256 neurons each, with the attention computed as usual \cite{vaswani2017attention,lee2018set}. 
The transformer is trained to minimize the L2 error of the next state prediction given the current state and action as inputs. 
The input is disentangled, and so has shape \texttt{(batch\_size, num\_components, num\_features)}. In each row (component representation) of each sample, features corresponding to other components are set to zero. 
The transformer is trained for 2000 steps with a batch size of 256, Adam optimizer \cite{kingma2014adam} with learning rate of 3e-4 and weight decay of 1e-5. 
For our reward function we use a fully-connected neural network with 1 hidden layer of 128 units. 
The reward network accepts an $(s, a, s')$ tuple as input (not disentangled) and outputs a softmax over the possible reward values of $[-1, 0, 1]$. 
It is trained for 2000 steps with a batch size of 512, Adam optimizer \cite{kingma2014adam} with learning reate of 1e-3 and weight decay of 1e-4. 
All hyperparameters were rather arbitrary (we used the default setting, or in case it did not work, the first setting that gave reasonable results, as was determined by inspection). 
To ensure that our model and reward functions are trained appropriately (i.e., do not diverge) for each seed, we confirm that the average loss of the CoDA model is below 0.005 at the training and that the average loss of the reward model is below 0.1, which values were found by inspection of a prototype run. These conditions were met by all seeds.

When used to produce masks, we chose a threshold of $\tau = 0.02$ by inspection, which seemed to produce reasonable results. A more principled approach would do cross-validation on the available data, as we did for Spriteworld (Appendix \ref{sec:inferring_local_factorization}). 

\paragraph{Tested configurations}

Table \ref{fig_batch_rl_and_fetch} reports a subset of our tested configurations. We report our results in full below. We considered the following configurations:
\begin{enumerate}
    \item \textbf{Real data only}.
    \item \textbf{CoDA + real data}: after training the CoDA model, we expand the base dataset by either 2, 3, 4 or 6 times.
    \item \textbf{Dyna (using CoDA model)}: after training the CoDA model, we use it as a forward dynamics model instead of for CoDA; we use 1-step rollouts with random actions from random states in the given dataset to expand the dataset by 2x. We also tried with 5-step rollouts, but found that this further hurt performance (not shown).  Note that Dyna results use only 5 seeds.
    \item \textbf{Dyna (using MBPO model)}: as the CoDA model exhibits significant model bias when used as a forward dynamics model, we replicate the state-of-the-art model-based architecture used by MBPO \cite{janner2019trust} and use it as a forward dynamics model for Dyna; we experimented with 1-step and 5-step rollouts with random actions from random states in the given dataset to expand the dataset by 2x. This time we found that the 5-step rollouts do better, which we attribute to the lower model bias together with the ability to create a more diverse dataset (1-step not shown). The MBPO model is described below. We use the same reward model as CoDA to relabel rewards for the MBPO model, which only predicts next state. 
    \item \textbf{MBPO + CoDA}: as MBPO improved performance over the baseline (real data only) at lower dataset sizes, we considered using MBPO together with CoDA. We use the base dataset to train the MBPO, CoDA, and reward models, as described above. We then use the MBPO model to expand the base dataset by 2x, as described above. We then use the CoDA model to expand \textit{the expanded dataset} by 3x the original dataset size. Thus the final dataset is 5x as large as the original dataset (1 real : 1 MBPO : 3 CoDA). 
\end{enumerate}

All configurations alter only the training dataset, and the same agent architecture/hyperparameters (reported above) are used in each case. 

\paragraph{MBPO model}

Since using the CoDA model for Dyna harms rather than helps, we consider using a stronger, state-of-the-art model-based approach. In particular, we adopt the model used by Model-Based Policy Optimization \cite{janner2019trust}. This model consists of a size 7 ensemble of neural networks, each with 4 layers of 200 neurons. We use ReLU activations, Adam optimizer \cite{kingma2014adam} with weight decay of 5e-5, and have each network output a the mean and (log) diagonal covariance of a multi-variate Gaussian. We train the networks with a maximum likelihood loss. To sample from the model, we choose an ensemble member uniformly at random and sample from its output distribution, as in \cite{janner2019trust}. 

\paragraph{Full results}

See Table \ref{table_appendix_batch_rl} for results.

\begin{adjustbox}{center,captionbelow={
Extended Batch RL results. 
Mean success ($\pm$ standard error, estimated using 1000 bootstrap resamples) on \texttt{Pong} environment. All results average over 10 seeds, except Dyna, which uses 5. 
},label=table_appendix_batch_rl,float=table}
\newcommand{\smol}[1]{{\scriptsize\texttt{#1}}}
\centering\small
\begin{tabular}{c@{\hskip 0.2in}ccc@{\hskip 0.16in}|@{\hskip 0.16in}ccccc}
	\toprule
$|\mathcal{D}|$	&Real data& Dyna & MBPO&
\multicolumn{5}{c}{Ratio of Real:CoDA [:MBPO] data (ours)}\\
$(1000s)$ &   1\smol{R}  &   1\smol{R}:1\smol{M} & 1\smol{R}:1\smol{M} &  1\smol{R}:1\smol{C} & 1\smol{R}:2\smol{C}   &  1\smol{R}:3\smol{C}  &  1\smol{R}:5\smol{C} & 1\smol{R}:3\smol{C}:1\smol{M} \\
	\midrule
$\hphantom{0}25$ & $13.2 \pm 0.7$  & $12.2 \pm 1.4$  & $18.5 \pm 1.5$  & $43.8 \pm 2.7$  & $41.6 \pm 3.7$  & $40.9 \pm 2.5$  & $38.4 \pm 4.9$  & $\mybm{46.8} \pm 3.2$  \\
$\hphantom{0}50$ & $22.8 \pm 3.2$  & $17.4 \pm 3.0$  & $36.6 \pm 4.0$  & $66.6 \pm 3.7$  & $58.4 \pm 4.7$  & $64.4 \pm 3.1$  & $62.5 \pm 3.5$  & $\mybm{70.4} \pm 3.7$  \\
$\hphantom{0}75$ & $43.2 \pm 4.7$  & $23.3 \pm 4.7$  & $46.0 \pm 4.5$  & $73.4 \pm 2.8$  & $71.6 \pm 3.8$  & $\mybm{76.7} \pm 2.6$  & $75.0 \pm 3.3$  & $74.6 \pm 3.3$  \\
$100$ & $63.0 \pm 3.0$  & $25.9 \pm 7.4$  & $66.4 \pm 5.2$  & $77.8 \pm 2.0$  & $77.4 \pm 1.8$  & $\mybm{82.7} \pm 1.5$  & $76.6 \pm 3.0$  & $73.7 \pm 2.9$  \\
$150$ & $77.4 \pm 1.3$  & $34.1 \pm 1.5$  & $72.6 \pm 5.6$  & $82.2 \pm 1.7$  & $84.0 \pm 1.2$  & $\mybm{85.8} \pm 1.4$  & $84.2 \pm 1.0$  & $79.7 \pm 3.5$  \\
$250$ & $78.2 \pm 2.7$  & $44.2 \pm 4.0$  & $77.9 \pm 2.3$  & $85.0 \pm 2.8$  & $\mybm{88.1} \pm 1.0$  & $87.8 \pm 1.7$  & $87.0 \pm 1.0$  & $78.3 \pm 5.0$  \\
	\bottomrule\vspace{-0.05in}
\end{tabular}
\end{adjustbox}

\subsection{Goal-conditioned RL}

Here we detail the procedure used in the Goal-conditioned RL experiments on the \texttt{Fetchpush-v1} and \texttt{Slide2} environments.

\paragraph{Implementation} This experiment uses the same codebase as our Batch RL, which provides state-of-the-art baseline HER agents and will be released with the paper.

\paragraph{Hyperparameters}
For \texttt{Fetchpush-v1} we use the default hyperparameters from the codebase, which outperform the original HER agents of \cite{andrychowicz2017hindsight,plappert2018multi} and follow-up works. We do not tune the CoDA agent (but see additional CoDA hyperparameters below). They are as follows:
\begin{itemize}
\item Off-policy algorithm: DDPG \cite{lillicrap2015continuous}
\item Hindsight relabeling strategy:  \texttt{futureactual\_2\_2} \cite{pitisprotoge}, using exclusively \texttt{future} \cite{andrychowicz2017hindsight} relabeling for the first 25,000 steps
\item Optimizer: Adam \cite{kingma2014adam} with default hyperparameters
\item Batch size: 2000
\item Optimization frequency: 1 optimization step every 2 environment steps after the 5000th environment step
\item Target network updates: update every 10 optimization steps with a Polyak averaging coefficient of 0.05
\item Discount factor: 0.98
\item Action l2 regularization: 0.01
\item Networks: 3x512 layer-normalized \cite{ba2016layer} hidden layers with ReLU activations
\item Target clipping: (-50, 0)
\item Action noise: 0.1 Gaussian noise
\item Epsilon exploration \cite{plappert2018multi}: 0.2, with an initial 100\% exploration period of 10,000 steps
\item Observation normalization: yes
\item Buffer size: 1M
\end{itemize}

On \texttt{Slide2} we tried to tune the baseline hyperparameters somewhat, but note that this is a fairly long experiment (10M timesteps) and so only a few settings were tested due to constraints. In particular, we considered the following modifications:
\begin{itemize}
\item Expanding the replay buffer to 2M (effective)
\item Reducing the batch size to 1000 (effective)* (used for results)
\item Using the \texttt{future\_4} strategy (agent fails to learn in 10M steps)
\item Reducing optimization step frequency to 1 step every 4 environment steps (about the same performance)
\end{itemize}

We tried similar adjustments to our CoDA agent, but found the default hyperparameters (used for results) performed well. We found that the CoDA agent outperforms the base HER agent on all tested settings.

For CoDA, we used the following additional  hyperparameters:
\begin{itemize}
\item CoDA buffer size: 3M
\item Make CoDA data every: 250 environment steps
\item Number of source pairs from replay buffer used to make CoDA data: 2000
\item Number of CoDA samples per source pair: 2
\item Maximum ratio of CoDA:Real data to train on: 3:1
\end{itemize}
\paragraph{Environment} 

\begin{wrapfigure}{r}{0.23\textwidth}
    \vspace{-\baselineskip}
	\centering
    \captionsetup{width=0.22\textwidth}
	\includegraphics[width=0.22\textwidth,height=0.2\textwidth]{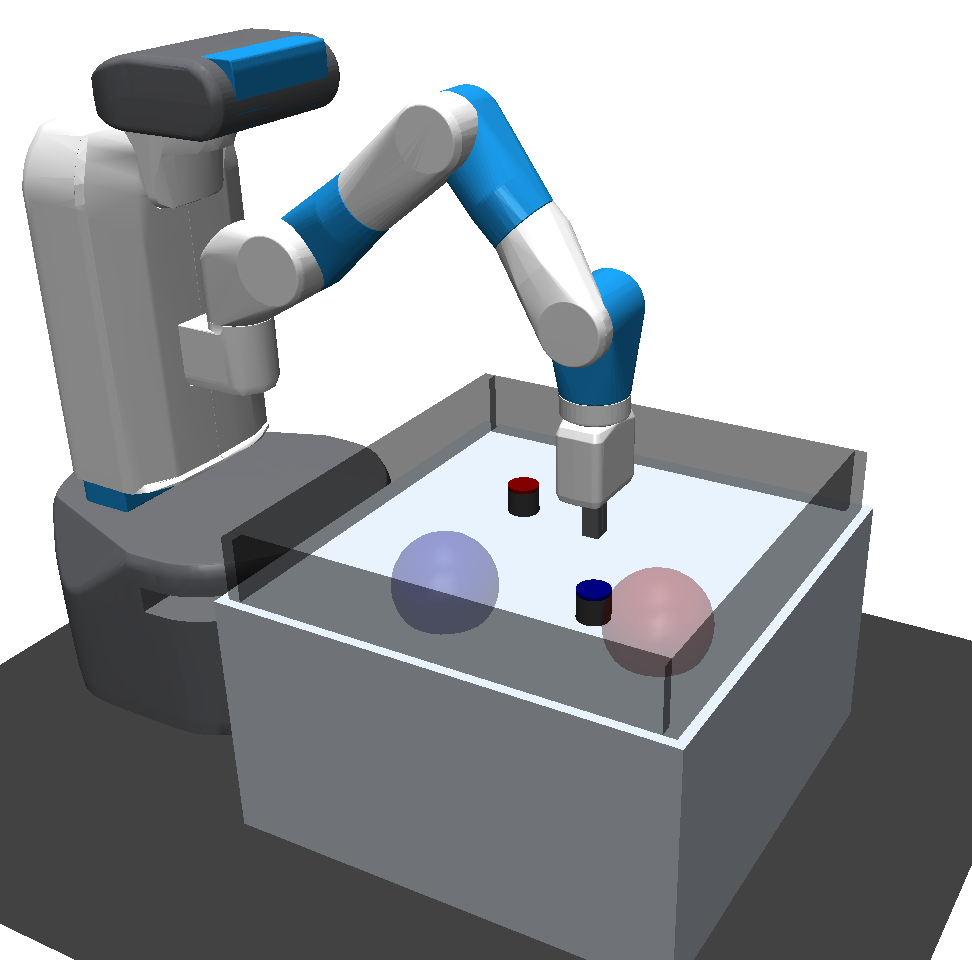}
	\caption{The \texttt{Slide2} environment.}
	\label{fig_env_fetch_appendix}
    \vspace{-\baselineskip}
\end{wrapfigure}

On \texttt{FetchPush-v1} the standard state features include the relative position of the object and gripper, which entangles the two. While this could be dealt with by dynamic relabeling (as used for HER's reward), we simply drop the corresponding features from the state. 

\texttt{Slide2} has two pucks that slide on a table and bounce off of a solid railing. Observations are 40-dimensional (including the 6-dimensional goal), and actions are 4-dimensional. Initial positions and goal positions are sampled randomly on the table. During training, the agent gets a sparse reward of 0 (otherwise -1) if \textit{both} pucks are within 5cm of their ordered target. At test time we count success as having both picks within 7.5cm of the target on the last step of the episode. Episodes last 75 steps and there is no done signal (this is intended as a continuous task).

\paragraph{CoDA Heuristic} For these experiments we use a hand-coded heuristic designed with domain knowledge. In particular, we assert that the action is always entangled with the gripper, and that gripper/action and objects (pucks or blocks) are disentangled whenever they are more than 10cm apart. This encodes independence due to physical separation, which we hypothesize is a very generally heuristic that humans implicitly rely on all the time.  The pucks have a radius of 2.5cm and height of 4cm, and the blocks are 5cm x 5cm x 5cm, so this heuristic is quite generous / suboptimal. Despite being suboptimal, it demonstrates the ease with which domain knowledge can be injected via the CoDA mask: we need only a high precision (low false positive rate) heuristic---the recall is not as important. It is likely that an agent could learn a better mask that also takes into account velocity.

\paragraph{Results plot} The plot shows mean $\pm 1$ standard deviation of the smoothed data over 5 seeds.

\section{Inferring Local Factorization}\label{sec:inferring_local_factorization}
Here we present several approaches to inferring the local factorization of subspaces, a crucial subroutine of CoDA.
We note that in many cases where domain knowledge is available, simple heuristics may suffice, e.g. in our Goal-conditioned RL experiments discussed in Section \ref{sec:experiments} where a simple distance-based indicator function in the state space was used. 
However, as such heuristics may not be universally available, the question of whether data-driven approaches can successfully infer local factorization is of general interest.
We note that the performance of CoDA will improve alongside future improvements in this inference task (motivating future work in this area), and that inferring the local factorization in general is an easier task than learning the environment dynamics.

We begin by presenting two methods for inferring local factorization, derived from variants of a next-state prediction task, which we here refer to as SANDy for Sparse Attention Neural Dynamics.
To verify the merits of SANDy in the local factorization inference task, we evaluate two SANDy variants in controlled settings where the ground truth factorization is known: first in a synthetic Markov process (MDP without actions), and second in Spriteworld.
This label information is used only to evaluate performance, and not to train the SANDy parameters.
The SANDy-Transformer model was used for the Online RL experiments presented in Section \ref{sec:experiments}.

\subsection{Methods}\label{subsection_learning_factorization}

We propose two Sparse Attention for Neural Dynamics (SANDy) models. In each case we seek to learn a function (or mask) $M(s, a) \to \{0, 1\}^{(|S|+|A|)\times|S|}$ whose output represents the adjacency matrix of the local causal graph, conditioned on the state and action. We note that $M(s, a)_{i,j} = 0$ alone is insufficient, in general, to determine the local subspace $\L \subset \S \times \A$, since there may be multiple disconnected subspaces $\L_k$ with $M_k(s, a)_{i,j} = 0$ whose union $\bigcup_k \L_k$ has $M_{\cup k}(s, a)_{i,j} = 1$. This can be resolved by our Proposition \ref{proposition_union_of_local_sets} if we also force the relevant structural equations to be the same. For now, we assume the mask determines the local subspace, and leave exploration of this possibility to future work. Empirically, we will see that our assumption is reasonable.

\paragraph{SANDy-Mixture} 

The first model is a mixture-of-MLPs model with an attention mechanism that is computed from the current state.
Each component of the mixture is a neural dynamics model with sparse local dependencies. 
For a given component, the key idea is to train a neural dynamics model to predict the next state $h(s_t, a_t) \approx s_{t+1}$ and approximate the masking function by thresholding the (transpose of the) network Jacobian of $h$, $[\mathbf{J}(s, a)]_{i, j} = \frac{\delta}{\delta [s, a]_j} [h(s, a)]_i$. Intuitively, we can think of $\mathbf{J}$ as providing the first-order element-wise dependencies between the predicted next state and the network input. We then derive the local factorization by thresholding the absolute Jacobian 
\begin{equation}
M_\tau(s, a) = \indicator{|\mathbf{J}(s, a)| > \tau},    
\label{eq:thresh}
\end{equation}
where $\indicator{\cdot}$ represents the indicator function and $\tau$ is a threshold hyperparameter.

To estimate the network Jacobian, we note that for standard activation functions (sigmoid, tanh, relu), it can be bound from above by the matrix product of its weight matrices. To see this, let $h_\theta$ be an $L$-layer MLP parameterized by $\theta = (\W^{(1)}, \B^{(1)}, \cdots, \W^{(L)}, \B^{(L)})$ with activation $\sigma$ with bounded derivative $\sigma'(x) \leq 1$, and note that for each layer $\bm{h}^{(\ell)} = \sigma(\W^{(\ell)}\bm{h}^{(\ell -1)} + \B^{(\ell)})$ we have:
\begin{align*}
	\Bigg\vert\frac{d\bm{h}^{(\ell)}_j}{d\bm{h}^{(\ell-1)}_i}\Bigg\vert &\leq \Big\vert\W^{(\ell)}_{ji}\Big\vert.
\end{align*}
Then, using the chain rule, triangle inequality, and the identity $|ab| = |a||b|$, we can compute:
\begin{align*}
	\Bigg\vert\frac{d\bm{h}^{(\ell)}_j}{d\bm{h}^{(\ell-2)}_i}\Bigg\vert &\leq \Bigg\vert\frac{d}{d\bm{h}^{(\ell-2)}_i}   \W^{(\ell)}_{j\cdot} \bm{h}^{(\ell-1)}\Bigg\vert =\Bigg\vert\W^{(\ell)}_{j\cdot} \cdot \frac{d\bm{h}^{(\ell-1)}}{d\bm{h}^{(\ell-2)}_i} \Bigg\vert \\
	&
	\leq \sum_k \Bigg\vert\W^{(\ell)}_{jk} \frac{d\bm{h}^{(\ell-1)}_k}{d\bm{h}^{(\ell-2)}_i} \Bigg\vert 
	\leq \Big\vert\W^{(\ell)}_{j\cdot} \Big\vert \cdot \Big\vert\W^{(\ell-1)}_{\cdot i} \Big\vert.
\end{align*}
Expanding this out to multiple layers, we see that $\big\vert \mathbf{J}_\theta(s, a)\big\vert  \leq \prod_{\ell \in [L]} |\W^{(\ell)}|$, as desired. We use this upper bound to approximate the network Jacobian of an MLP by setting $\mathbf{\hat J} = \prod_{\ell \in [L]} |\W^{(\ell)}|$. A similar idea is used by \cite{germain2015made,lachapelle2019gradient}, among others, to control element-wise input-output dependencies.

We use this static approximation $\mathbf{\hat J}$ to facilitate learning a sparse dynamic mask by specifying a mixture model $h(s, a) = \sum_i \alpha^{(i)}_\phi(s,a) h^{(i)}_\theta(s,a)$ over the environment dynamics (with $\sum_i \alpha^{(i)}_\phi(s,a) = 1 \ \forall \ (s, a)$), where each component $h^{(i)}_\theta$ is an MLP as specified above with a sparsity prior on its Jacobian bound $\mathbf{\hat J}^{(i)}$ to encourage sparse (i.e. well-factorized) local solutions. The dynamic mask is computed by first approximating the Jacobian, $\mathbf{\hat J}(s,a) = \sum_i \alpha^{(i)}_\phi(s,a)\mathbf{\hat J}^{(i)}$, then thresholding by $\tau$ as in Equation \ref{eq:thresh}.

Note that we assume the mixture components $\alpha$ are a function of the current state and action alone; in other words the factorization (captured by the network Jacobian of each component) can be \emph{locally inferred}.
The next-state prediction is given by $\hat \s_{t+1} = ( h_\theta^{(1)}(\s_t, \mathbf{a}_t), \cdots, h_\theta^{(N)}(\s_t, \mathbf{a}_t))^T \bm{\alpha}_\phi(\s_t, \mathbf{a}_t)$. To train the model, we optimize the objective:
\begin{equation}\label{eq:mixture-mlp}\small
\underset{\theta, \phi}{\minimize} \ 
  \frac{1}{|\mathcal{D}|} \Big( \sum_{(\s_t, \mathbf{a}_t \s_{t+1}) \in \mathcal{D}} ||\s_{t+1} - h(\s_t, \mathbf{a}_t)||_2^2 \Big) 
  + \lambda_1 S(\theta) +  \lambda_2 R(\phi) + \lambda_3 ||(\theta, \phi)||_2
\end{equation}
where $S(\theta) = \frac{1}{K} \sum_i |\hat{\mathbf{J}}_\theta^{(i)}|_1$ puts an $\ell_1$ prior on each mixture component to induce sparsity,
and $R(\phi)$ encourages high entropy in the attention probabilities
\[
  R(\phi) = \frac{1}{|\mathcal{D}|} \sum_{\s \in \mathcal{D}} \sqrt{\frac{1}{N}\sum_{j \in [N]} [A_\phi(\s, \mathbf{a})]_j}.
\]

We note that more sophisticated methods of computing Jacobians of neural networks--including architectural changes and optimization strategies \cite{arjovsky2017wasserstein}---have been proposed, and could in principle be used here as well.

\paragraph{SANDy-Transformer}

As an alternative model, we use a transformer-like architecture that applies self-attention between a set of (potentially multi-dimensional) inputs \cite{vaswani2017attention, lee2018set}. Our architecture is composed of a stacked self-attention blocks. Each block accepts a set of inputs $X = \{x_i \in \R^n\}$ and composes three functions of each input: query $Q: \R^n \to \mathbb{R}^d$, key $K: \R^n \to \mathbb{R}^d$, and value $V: \R^n \to \mathbb{R}^m$. The block returns a set of outputs $\{y_i \in \mathbb{R}^m\}$ of size $|X|$, each of which is computed as: $y_i = A_i^TV(X)$, where $A_i = \textrm{softmax}\big(\sum_j(Q(x_i)_j K(x_j)_i)\big)$ (note that $V(X)$ is a matrix of size $|X| \times m$). We approximate the block mask (approximate Jacobian) as $A = [A_1, A_2, \dots, A_{|X|}] \in \R^{|X| \times |X|}$, and the full network mask as the product of the block masks (as in the SANDy-Mixture). We used two-layer MLPs for each function $K, Q, V$ in Spriteworld and three-layer MLPs in Pong. 

To apply this architecture to our problem, we first embed each state and action component (single feature, or group of features) into $\R^n$ to produce a set of inputs $X$ and pass this through each stacked self-attention block to obtain a set of outputs $Y$. We then discard any output components that correspond to the action features to obtain the next state prediction and mask. The network $h$ is trained to minimize mean squared error:
\begin{equation}\label{eq:sandy-transformer}\small
\underset{\theta}{\minimize} \ 
  \frac{1}{|\mathcal{D}|}\sum_{(\s_t, \mathbf{a}_t \s_{t+1}) \in \mathcal{D}} ||\s_{t+1} - h(\s_t, \mathbf{a}_t)||_2^2.
\end{equation}
Unlike the SANDy-Mixture, no sparsity regularizers are applied, as we found the sparsity induced by the softmax attention mechanism to be sufficient. 

\subsection{Evaluation environments}

\paragraph{Synthetic Markov Processes}
We investigate the capacity of the SANDy-Mixture to learn simple factorized transition dynamics under a globally factored Markov Process (\textsc{StationaryMP}).
Unlike the general MDP setting, no agent/policy is considered.
However the ability to train an unconstrained dynamics model to approximate factorized environment dynamics is an important subtask within our overall approach.
Assuming a spherical Gaussian prior over the initial states $\s^0 \in \R^9$, the \textsc{StationaryMP} is entirely specified by transition distribution $p(\s^{t+1}|\s^t)$.
We assume that the state transitions factorize (globally) into three parts.
Denoting by $\s_{n \cdots m}^t$ the $n$-th through $m$-th dimensions of the time $t$ state $\s^t$, we have:
\[
p(\s^{t+1}_{1\cdots9} | \s^{t}_{1\cdots9}) =
  p(\s^{t+1}_{1\cdots4} | \s^{t}_{1\cdots4})
  p(\s^{t+1}_{5\cdots7} | \s^{t}_{5\cdots7})
  p(\s^{t+1}_{8,9} | \s^{t}_{8,9}).
\]
In other words, we have a block-diagonal transition matrix comprising three blocks with sizes $4$, $3$, and $2$, respectively.
In our case, all transition factors are deterministic non-linear mappings, e.g. $p(\s^{t+1}_{1\cdots4} | \s^{t}_{1\cdots4}) = \delta(g_{1\cdots4}(\s^{t}_{1\cdots4}))$, with $g_{1\cdots4}: \R^{4} \rightarrow \R^{4}$ is a randomly-initialized single-hidden-layer neural network with $32$ hidden units and GELU nonlinearity \cite{hendrycks2016gaussian}.
In this case, we can alternatively express the deterministic dynamics via the transition function
\begin{align}\label{eq:StationaryMP}
\s^{t+1} &= (\s_{1\cdots4}^{t+1}, \s_{5\cdots7}^{t+1}, \s_{8, 9}^{t+1}) \nonumber\\ 
  &= (g_{1\cdots4}(\s_{1\cdots4}^{t}), g_{5\cdots7}(\s_{5\cdots7}^{t}), g_{8,9}(\s_{8,9}^{t})).
  \tag{\textsc{StationaryMP}}
\end{align}

We now turn to a more sophisticated MP with locally factored dynamics (\textsc{$\epsilon$-NonstationaryMP}), and investigate whether the SANDy-Mixture can learn to recognize local disentanglement. 

We refer to the component functions ${g_{1\cdots4}: \R^4 \rightarrow \R^4}$, ${g_{5\cdots7}: \R^3 \rightarrow \R^3}$, and ${g_{8,9}: \R^2 \rightarrow \R^2}$ used in the \ref{eq:StationaryMP} as \emph{local} transitions.
We now introduce \emph{global} interactions via the global transition functions, ${G_{1\cdots4}: \R^4 \rightarrow \R^9}$, ${G_{5\cdots7}: \R^4 \rightarrow \R^9}$, and ${G_{8, 9}: \R^4 \rightarrow \R^9}$, which respectively map the local state factors onto the global state space\footnote{
Like the local transition functions, global transition functions are implemented as randomly-initialized single-hidden-layer neural networks.
}.
This allows us to extend the \ref{eq:StationaryMP} by adding global state transitions to the local state transitions whenever the norm of the local state factors exceeds the value of a hyperparameter $\epsilon$ (lower $\epsilon$ indicates more global interaction).
Denoting by $\indicator{\cdot}$ the indicator function, we have
\begin{align*}\label{eq:NonstationaryMP}
\s^{t+1} &= \left[(\s_1^{t+1}, \s_2^{t+1}, \s_3^{t+1}, \s_4^{t+1}), (\s_5^{t+1}, \s_6^{t+1}, \s_7^{t+1}), (\s_8^{t+1}, \s_9^{t+1}) \right]\\
&= \left[g_{1\cdots4}(\s_{1\cdots4}^{t}),  g_{5\cdots7}(\s_{5\cdots7}^{t}),  g_{8,9}(\s_{8,9}^{t})\right]\\
  &\quad + G_{1\cdots4}(\s_{1\cdots4}^{t}){\indicator{||\s_{1\cdots4}^{t}||_2 > \epsilon}} \nonumber \\
  &\quad + G_{5\cdots7}(\s_{5\cdots7}^{t}){\indicator{||\s_{5\cdots7}^{t}||_2 > \epsilon}} \nonumber \\  
  &\quad + G_{8,9}(\s_{8,9}^{t}){\indicator{||\s_{8,9}^{t}||_2 > \epsilon}}.
  \tag{\textsc{$\epsilon$-NonstationaryMP}}
\end{align*}

\paragraph{Spriteworld}

Since we extended \texttt{Spriteworld} with a ground truth mask renderer, we are able to directly evaluate our SANDy models in \texttt{Spriteworld} as well. See the main text and Appendix \ref{sec:training_details} for a description of the \texttt{Spriteworld} environment. 

\subsection{Results}\label{sec:mask_results}

In this Subsection we measure ability of the proposed SANDy algorithm (in its two variants) to correctly infer local factorization.
At each transition we can query the environment for the ground-truth connectivity pattern of the local causal graph: given $|S|+|A|$ dimensions of current state and action and $|S|$ dimensions of next state, this corresponds an adjacency matrix $\mathbf{Y} \in \{0, 1\}^{|S|+|A|\times|S|}$.
We note that accessing these evaluation labels---which are not used to train SANDy---requires a controlled synthetic environment like the ones we consider, and we leave design of an evaluation protocol suitable for real-world environments to future work.

We learn the SANDy network parameters using a training dataset of $40,000$ transitions, with an additional validation dataset of $10,000$ transitions used for early stopping and hyperparameter selection.
We used the Adam optimizer with learning rate of $0.001$ and default hyperparameters.
In the \textsc{$\epsilon$-NonstationaryMP}, we set $\epsilon=1.5$, while in the Spriteworld setting we collect training trajectories by deploying a random-action agent in the environment, and randomly resetting the environment with 5\% probability at every step to increase diversity of experiences.
We then evaluate the SANDy models by computing local factorizations $M_\tau(s, a): |S| \times |A| \to \{0, 1\}^{(|S|+|A|)\times|S|}$ as a function of the threshold $\tau$ for each transition in a held-out test dataset of $10,000$ trajectories.
We compute true and false positive rates for various values of $\tau$ to produce ROC plots.

\begin{figure}[th]\small
\centering
\begin{subfigure}{.3\textwidth}
  \includegraphics[width=\textwidth]{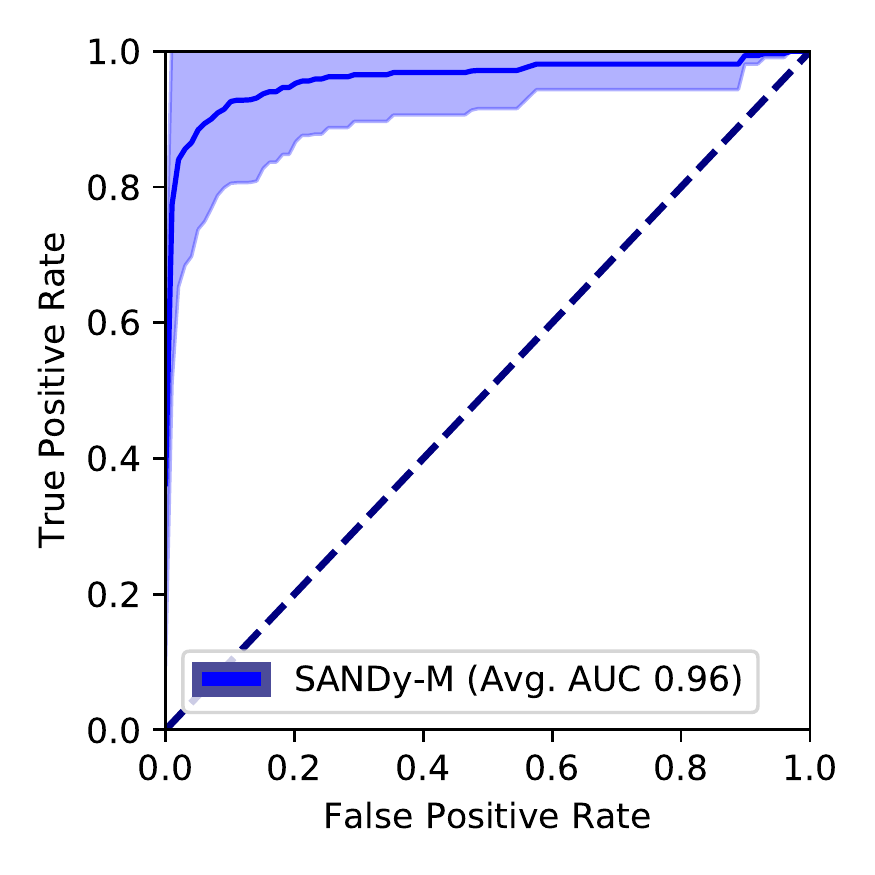}
  \caption{\textsc{StationaryMP}}
  \end{subfigure}%
\hfill
\begin{subfigure}{.3\textwidth}
  \includegraphics[width=\textwidth]{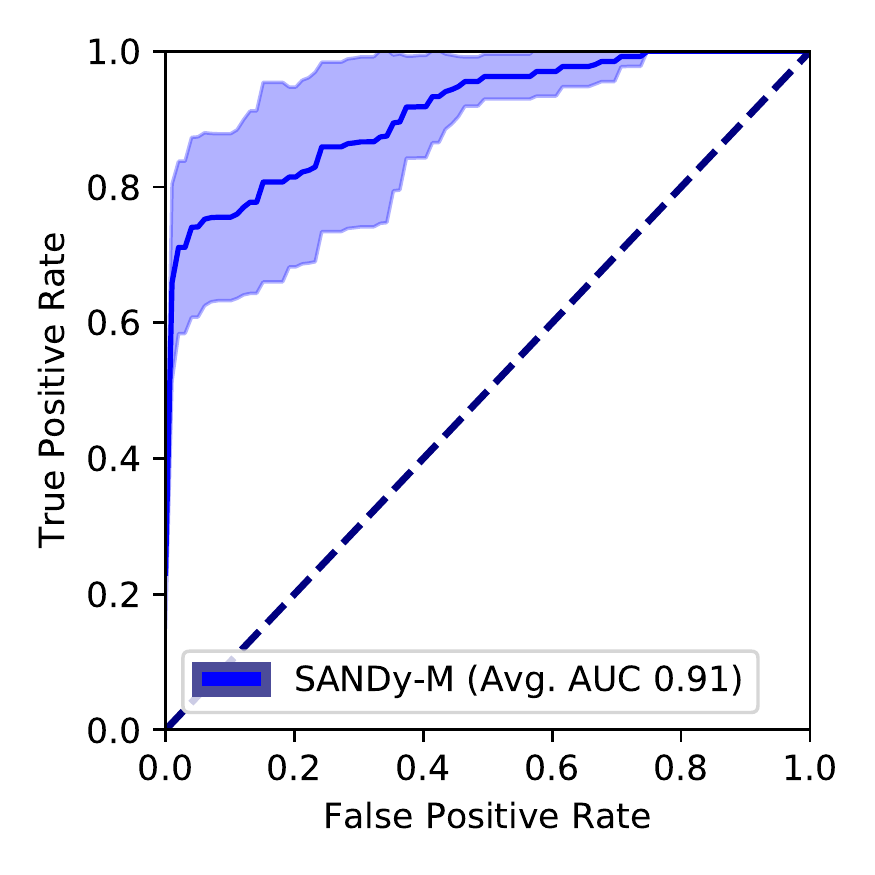}
  \caption{\textsc{$\epsilon$-NonstationaryMP}}
  \end{subfigure}%
\hfill
\begin{subfigure}{.3\textwidth}
  \includegraphics[width=\textwidth]{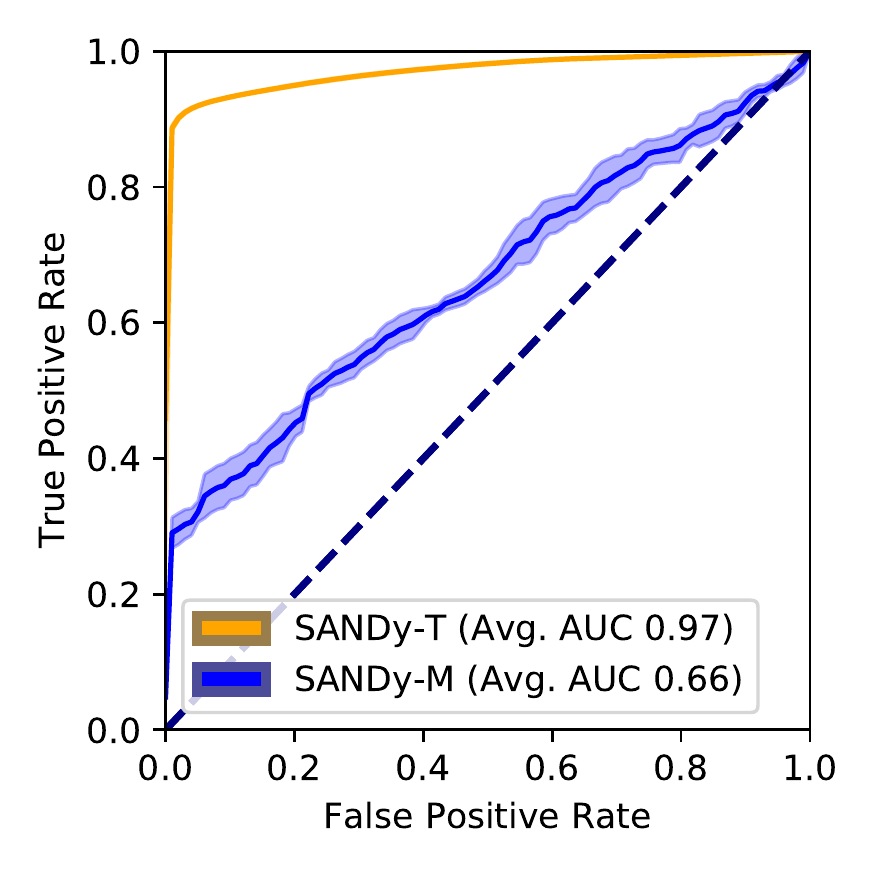}
  \caption{\texttt{Spriteworld}}
\end{subfigure}%
\caption{ 
ROC plots for correct sparsity patter prediction on the three environments.
On held-out test transitions we derive the ground truth local connectivity per step---label information that is \emph{not} used to train the attention model---and measure (over 5 runs; $1$ std. dev. shaded) true and false positive rates while sweeping the mask threshold $\tau$ over its allowable range.
An accurate model generates an Area Under the Curve (AUC) close to $1$.
We observe that while SANDy-Mixture is sufficient for (nearly) solving the simpler synthetic MP environments, it underperforms in Spriteworld.
SANDy-Transfomer, which has a stronger inductive bias, is sufficient to (nearly) solve Spriteworld.
}
\label{fig:roc}
\end{figure}

Figure \ref{fig:roc} shows that while the SANDy-Mixture is sufficient to solve the simpler synthetic MP settings (with avg AUC of $0.96$ and $0.91$ for the stationary and non-stationary variants), it scales poorly to the Spriteworld environment.
While the modest inductive bias of sparse local connections and high-entropy mixture components in SANDy-Mixture makes it widely applicable, we hypothesize that its sensitivity to hyperparameters makes it difficult to tune in complex settings.
Fortunately, SANDy-Transformer, performs favorably in Spriteworld by incorporating a stronger inductive bias about the state subspace structure.
Thus we use SANDy-Transformer to perform local factorization inference in the remaining experiments.

\begin{figure}[ht]
\centering
\subcaptionbox{SANDy-Mixture}{
\includegraphics[width=0.4\textwidth]{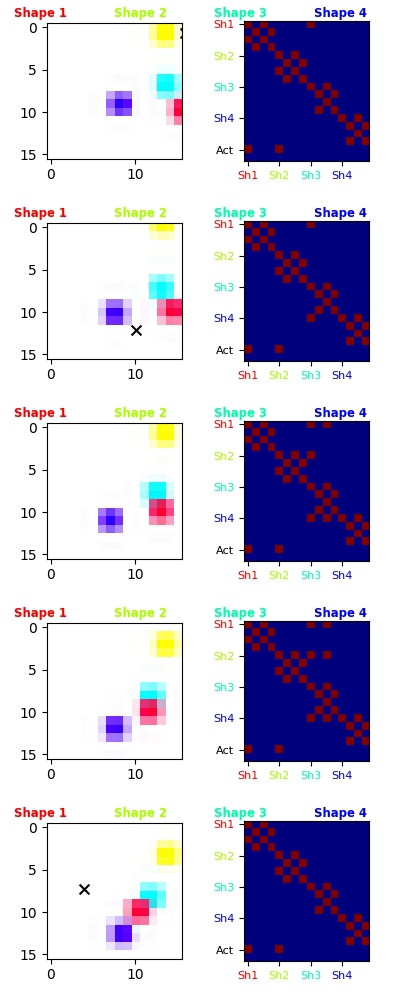}
\label{fig:qual_attn_left}
}
\subcaptionbox{SANDy-Transformer}{
\includegraphics[width=0.4\textwidth]{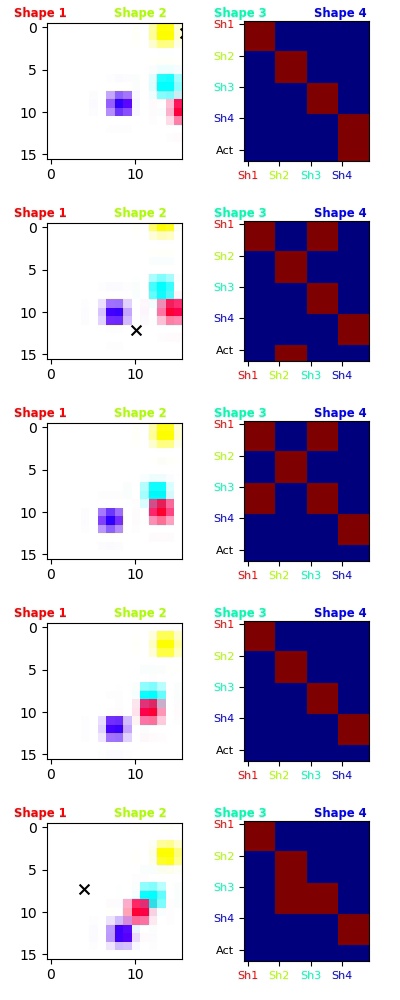}
\label{fig:qual_attn_right}
}
\caption{
Qualitative comparison of two attention mechanisms on the same \texttt{Spriteworld} trajectory.
\textbf{SANDy-Mixture} (left) has a weaker inductive bias as it relies on sparsity regularization alone.
Accordingly, it can learn a more compact subspace (e.g. grid patterns within a shape indicate that $x$ and $y$ coordinates move independently), but is less reliable in attending to collisions between shapes, and completely fails to attend to the action's affect on shapes.
\textbf{SANDy-Transformer} has a stronger inductive bias and can more reliably infer the local interaction pattern between the five subspaces (four shapes and one action).
}
\label{fig:qual_attn}
\end{figure}

Figure \ref{fig:qual_attn} provides some qualitative intuition as to how the two variants of SANDy differ in their attention strategy in the Spriteworld environment.

\section{Fitting dynamics models to Spriteworld}\label{appdx_dynamics_modeling}

\begin{figure}[!t]
\centering
\subcaptionbox{Ground Truth}{
\includegraphics[width=1.0\textwidth]{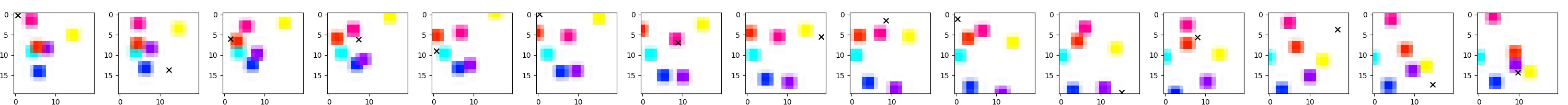}
\label{fig:qual_dynamics_models_1}
}
\subcaptionbox{Linear Dynamics}{
\includegraphics[width=1.0\textwidth]{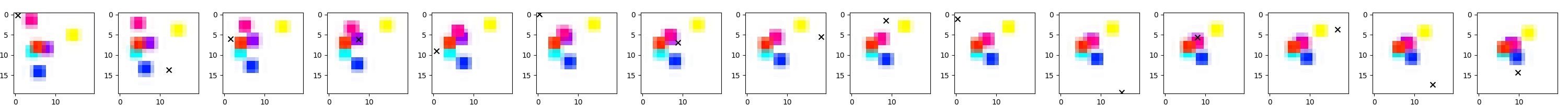}
\label{fig:qual_dynamics_models_2}
}
\subcaptionbox{MLP Dynamics}{
\includegraphics[width=1.0\textwidth]{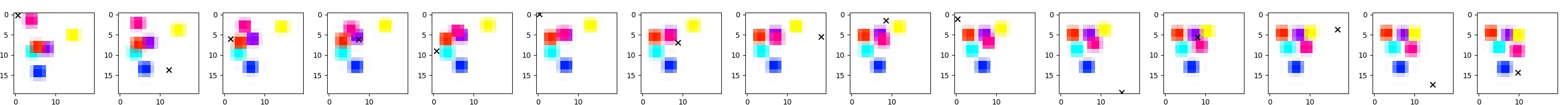}
\label{fig:qual_dynamics_models_3}
}
\subcaptionbox{LSTM Dynamics}{
\includegraphics[width=1.0\textwidth]{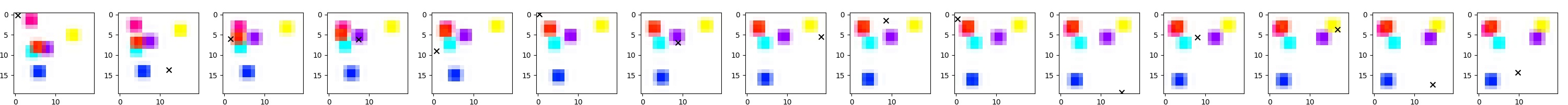}
\label{fig:qual_dynamics_models_4}
}
\caption{
Auto-regressive model-based rollouts for a variety of dynamics models fit to Spriteworld.
While the dynamics models were able to achieve relatively low error in the next-state prediction task, they fail to capture collisions and long-term dependencies in the sampled trajectories, and thus were omitted as baselines in the RL experiments.
All trajectories share the same initial state.
}
\label{fig:qual_dynamics_models}
\end{figure}

Since CoDA is a data augmentation strategy, it is reasonable to consider an alternative approach to augmenting the experience buffer: sampling from a dynamics model as in model-based RL.
Here we present some qualitative results from our efforts in fitting dynamics models to the Spriteworld environment.
We found that while dynamics models achieve a decent error in the next-state prediction task, they fail to produce a diverse set of trajectories when used as autoregressive samplers.
In particular, the autoregressive sampling did not model collisions well and often produced trajectories where sprites converged to fixed points in space after a short number of steps.
Figure \ref{fig:qual_dynamics_models} shows trajectories sampled autoregressively from Linear, MLP, and LSTM-based dynamics models, alongside the ground truth trajectory.
Note that all dynamics models were trained to minimize error in next-state prediction given the current state and action. 
In other words the LSTM auto-regressively predicts successive dimensions of the next state rather than modeling multiple time steps of the trajectory.
Nevertheless the environment is truly Markov because instantaneous velocities are observed, so this information should be sufficient in theory to capture the environment dynamics.

\section{Sample efficient dynamics modeling with CoDA}\label{appdx_coda_for_mbrl}

\begin{figure}[h!]
\centering
\includegraphics[width=0.5\textwidth]{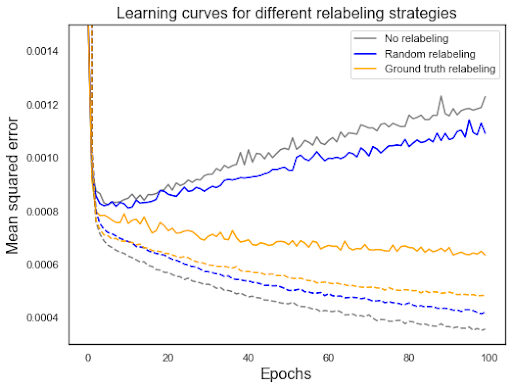}
\caption{
Learning curves for training a forward dynamics model using data from a random policy. Dotted lines indicate training performance, whereas solid lines indicate validation performance. We see that using the ground truth mask prevents overfitting and allows us to achieve much better validation performance. 
}
\label{fig:training_forward_model}
\end{figure}

If we had access to the ground truth local factorization, even for a few samples (e.g., we could have humans label them), how much more efficient would it be to train a dynamics model? In Figure \ref{fig:training_forward_model}, we sample 2000 transitions from a random policy in Spriteworld and use the data to train a forward dynamics model using MSE loss. The validation loss throughout training is plotted. Our baseline uses only the initial dataset, and quickly overfits the training set, showing increasing error after the initial few epochs. The same applies to a ``random'' CoDA strategy, that does CoDA using an identity attention mask ($M(s, a) = I \ \forall \ (s, a)$) to randomly relabel the components. The random strategy does a bit better than the no CoDA strategy, since the randomness acts as a regularizer. Finally, we train a model using an additional 35,000 unique counterfactual CoDA transitions, and find that it significantly improves validation loss and prevents the model from overfitting. Note that we could have generated many more CoDA samples: from 2000 base transitions, if 80\% of them do not involve collisions and there are 4 connected components in each, we could generate as many as $1600^4$ (6.5 trillion!) counterfactual samples.  

\section{Compute Infrastructure}

Experiments were run on a mix of local machines and a compute cluster, with a mix of GTX 1080 Ti, Titan XP, and Tesla P100 GPUs. This was solely to run jobs in parallel, and all experiments can be run locally (GPU optional for \texttt{Spriteworld} and \texttt{Pong}, but recommended for \texttt{Fetch} experiments).

\end{document}